%% file: iccc.tex
\documentclass[letterpaper]{article}
\usepackage{iccc}

\usepackage{times}
\usepackage{helvet}
\usepackage{courier}

\usepackage{graphicx}
\usepackage{caption}
\usepackage{subcaption}
\usepackage[dvipsnames]{xcolor}
\usepackage{tcolorbox}
\usepackage{booktabs}
\usepackage{multirow}
\usepackage{soul}

\newtcolorbox{highlight}{
  colback=lightgray!20, 
  colframe=gray!60, 
  boxrule=1pt, 
  arc=0pt, 
  boxsep=0pt, 
  left=6pt, 
  right=6pt, 
  top=6pt, 
  bottom=6pt, 
}

\newtcolorbox{highlightgreen}{
  colback=green!20, 
  colframe=green!60, 
  boxrule=1pt, 
  arc=0pt, 
  boxsep=0pt, 
  left=6pt, 
  right=6pt, 
  top=6pt, 
  bottom=6pt, 
}

\newcommand{\redhl}[1]{{#1}}
\newcommand{\greenhl}[1]{{#1}}
\newcommand{\apricothl}[1]{{#1}}
\newcommand{\skybluehl}[1]{{#1}}
\newcommand{\redviolethl}[1]{{#1}}
\newcommand{\orchidhl}[1]{{#1}}
\newcommand{\melonhl}[1]{{#1}}

\title{Evaluating Creative Short Story Generation \\in Humans and Large Language Models}

\author{
 \textbf{Mete Ismayilzada\textsuperscript{1,2,4}},
 \textbf{Claire Stevenson\textsuperscript{3}},
 \textbf{Lonneke van der Plas\textsuperscript{2,4}}
\\
 \textsuperscript{1}EPFL,
 \textsuperscript{2}Idiap Research Institute,
 \textsuperscript{3}University of Amsterdam,\\
 \textsuperscript{4}Università della Svizzera Italiana
\\
 \small{
   mahammad.ismayilzada@epfl.ch
 }
}

\begin{document} 
\maketitle
\begin{abstract}
\begin{quote}
\input{00_abstract}
\end{quote}
\end{abstract}

\input{01_introduction}
\input{02_related_work}
\input{03_methodology}
\input{04_results}
\input{06_discussion}
\input{08_limitations}
\input{09_ethics}


\input{10_acknowledgements} 

\newpage
\bibliographystyle{iccc}
\bibliography{iccc}

\newpage
\appendix

\input{11_appendix}

\end{document}

%% file: 00_abstract.tex
Story-writing is a fundamental aspect of human imagination, relying heavily on creativity to produce narratives that are novel, effective, and surprising. While large language models (LLMs) have demonstrated the ability to generate high-quality stories, their creative story-writing capabilities remain under-explored. In this work, we conduct a systematic analysis of creativity in short story generation across 60 LLMs and 60 people using a five-sentence cue-word-based creative story-writing task. We use measures to automatically evaluate model- and human-generated stories across several dimensions of creativity, including novelty, surprise, diversity, and linguistic complexity. We also collect creativity ratings and Turing Test classifications from non-expert and expert human raters and LLMs. Automated metrics show that LLMs generate stylistically complex stories, but tend to fall short in terms of novelty, surprise and diversity when compared to average human writers. Expert ratings generally coincide with automated metrics. However, LLMs and non-experts rate LLM stories to be more creative than human-generated stories. We discuss why and how these differences in ratings occur, and their implications for both human and artificial creativity.

%% file: 01_introduction.tex
\section{Introduction}
\label{sec:introduction}
\begin{figure}[t]
    \centering   
    \includegraphics[width=\linewidth,trim={1cm 1.5cm 0cm 0cm},clip]{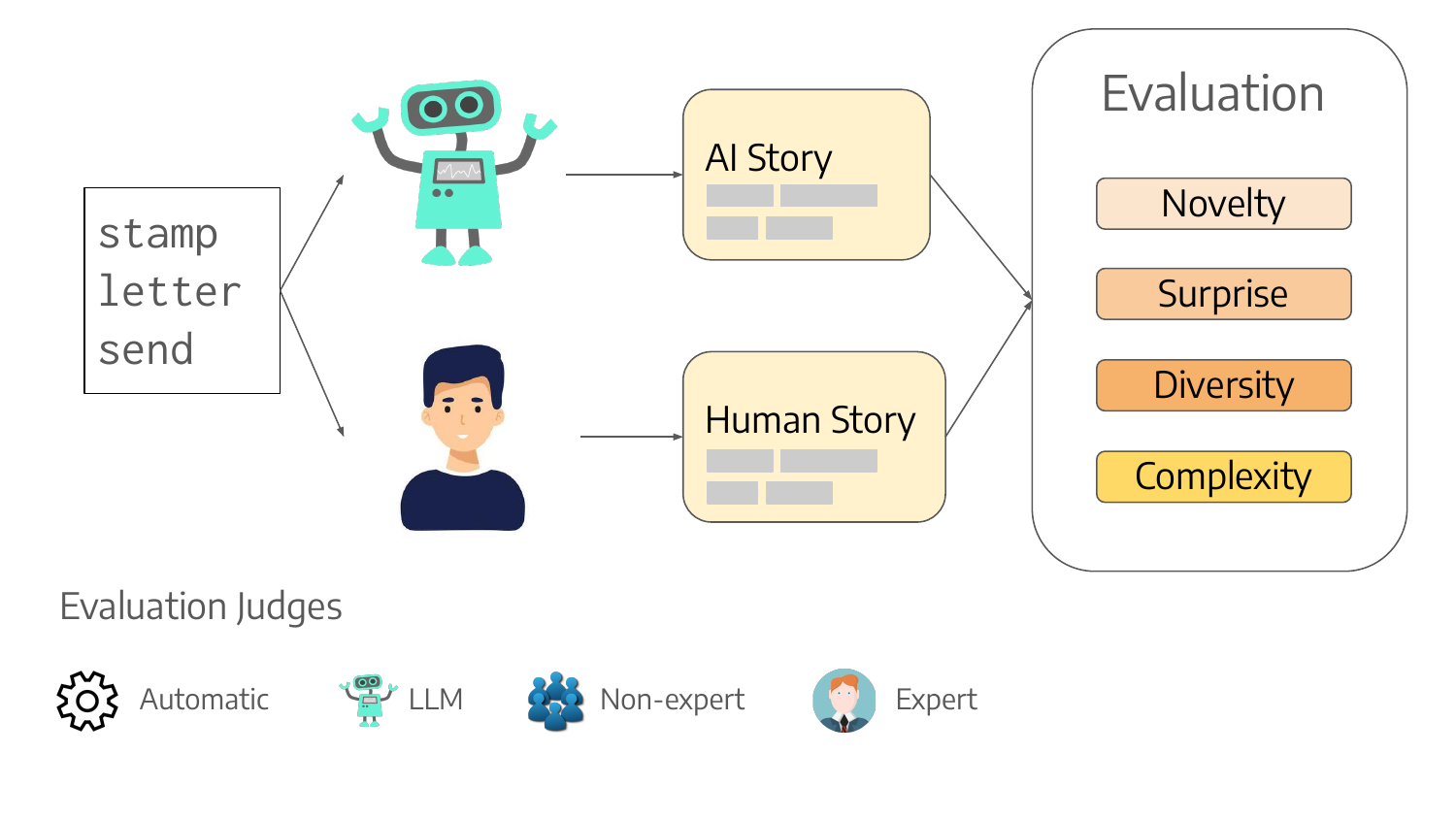}
    \caption{Our study setup illustrated with an example. Both humans and LLMs are asked to write a creative short story using three cue words and evaluated on several creativity metrics by human (experts vs non-experts) and LLM judges.}
    \label{fig:main-figure}
\end{figure} 
Story-writing lies at the core of human imagination and communication, serving as a potent means to connect and convey ideas effectively spanning across all human cultures and time periods \cite{Barthes1975AnIT}. It typically demands creativity, especially when shaping a captivating and persuasive narrative. Creativity is the ability to produce novel, useful, and surprising ideas, and has been widely studied as a crucial aspect of human cognition \cite{Boden1991TheCM,guilford1967nature,barron1955disposition,stein1953creativity}. While humans are natural storytellers, getting machines to generate stories automatically has been a long-time challenge \cite{ang2011theme,zhu2010towards,gervas2010story,Meehan1977TALESPINAI,lebowitz1984creating}. However, recently large language models \cite{zhao2024surveylargelanguagemodels} have been shown to produce high-quality short and long stories on arbitrary topics \cite{yang-etal-2022-re3,goldfarb-tarrant-etal-2020-content}. These stories are often evaluated by humans on their global coherence, relevance to the premise, repetitiveness, and general interestingness to the reader \cite{chhun-etal-2022-human}. However, the extent to which these model-generated stories are truly creative, i.e. \textit{novel}, \textit{effective}, and \textit{surprising}, remains under-studied. 

LLM creativity has generally been evaluated with tasks requiring short responses such as words or phrases. For example, many works have employed the Alternative Uses Test \cite{guilford1967nature}, where people and models are asked to come up with creative uses for an everyday object like a \textit{brick} and reported near-human performance results \cite{stevenson2022putting,GesPushingGC,Hubert2024TheCS,Koivisto2023BestHS,gilhooly2023ai}. However, the extent to which these results generalize to creativity tasks requiring longer and more complex responses remains underexplored.

Recent works evaluating LLM's ability to produce creative content have shown that models largely fall behind professional human writers \cite{Tian2024AreLL,Marco2024SmallLM,Marco2024PronVP,Chakrabarty2023ArtOA}. On the other hand, Orwig et al. (\citeyear{orwig2024language}) finds no significant difference between average human and AI-generated short stories in terms of creativity ratings by non-experts or GPT-4, when comparing humans to ChatGPT models. The extent to which these results hold when considering a collection of different models and evaluations across multiple dimensions of creativity as well as expert and non-expert raters remains unclear.


To bridge these gaps, in this work, we conduct a systematic analysis of creativity in short story generation in humans and LLMs. We employ a creative short story generation task that is typically used in psychology to measure the creativity of humans \cite{prabhakaran2014thin,johnson2023divergent,orwig2024language}. In this task, the goal is to write a short creative story in approximately five sentences based on three cue words such as \textit{stamp}, \textit{letter} and \textit{send}. We evaluate 60 humans and 60 state-of-the-art instruction-finetuned large language models on this task and analyze their performance based on multiple automatic metrics of creativity representing, overall creativity, and the specific aspects of novelty, surprise, diversity, and complexity. Our analysis shows that model-generated stories tend to employ more complex linguistic structures than humans; however, they significantly fall short when it comes to novelty, diversity, and surprise compared to average human writers. 

Additionally, we collect fine-grained creativity judgments from non-expert and expert human raters and LLMs for both human and model stories. We find that while non-expert raters and LLMs rate LLM-generated stories as more creative than human-generated stories, expert judgments positively correlate with automated metric results. Our further analysis shows that non-expert human and LLM ratings are driven by linguistic complexity of the stories (e.g. number of words) while expert raters focus on the semantic complexity. Similarly, we find that experts are much more reliable in distinguishing between human-generated and AI-generated stories. Finally, we discuss the implications of our work for both human and machine creativity.

%% file: 02_related_work.tex
\section{Related Work}
\label{sec:related_work}

\paragraph{Creativity Evaluation}
Evaluating creativity is a challenging task due to its subjective nature, however, several evaluation methods have been proposed in the past \cite{lamb2018evaluating,amabile1982social}. The most common method of creativity evaluation is called the Consensual Assessment Technique (CAT) \cite{amabile1982social}. CAT relies on the collective judgment of human experts. However, the level of expertise required for a human rater is a subject of debate, with most evidence favoring (quasi-)experts over non-experts as good raters \cite{kaufman2012beyond,hu2023comparing,long2022dissecting,ceh2022assessing,veale2015game,lamb2015human,gervas2019exploring}. With the rise of powerful generative AI models, LLMs are increasingly being used as judges in many evaluation tasks including creativity \cite{bavaresco2024llms,chen2024humans,Liu2023GEvalNE,Gilardi2023ChatGPTOC,Chiang2023CanLL,organisciak2023beyond}. Consequently, in addition to automated metrics, we also conduct evaluations with human expert and non-expert raters and LLMs and study similarities and differences between these different sources of judgment.

Other creativity evaluation methods are generally theoretical frameworks that aim to comprehensively evaluate creativity \cite{lamb2018evaluating} or manual psychometric tests that call for brief responses, such as single words or short phrases \cite{guilford1967nature,torrance1966torrance}. LLM creativity has also predominantly been assessed using these psychometric tasks. For instance, numerous studies have utilized the Alternative Uses Test \cite{guilford1967nature}, in which participants, both human and AI models, generate creative uses for everyday objects like a brick, often showing near-human performance \cite{stevenson2022putting,GesPushingGC,Hubert2024TheCS,Koivisto2023BestHS,gilhooly2023ai}. In this work, we instead focus on evaluating creativity of humans and LLMs in story generation task, which requires longer and complex responses.


\paragraph{Creative Story Evaluation}
While most works have focused on evaluating model-generated stories on global coherence, relevance to premise, repetitiveness, and overall interestingness \cite{chhun-etal-2022-human}, recent studies have also evaluated the creativity of AI models in producing stories \cite{Tian2024AreLL,orwig2024language,johnson2023divergent,Marco2024SmallLM,Marco2024PronVP,Chakrabarty2023ArtOA}. Chakrabarty et al. (\citeyear{Chakrabarty2023ArtOA}) generates short stories using LLMs based on plots from popular fictional works featured in the New York Times and performs a detailed expert evaluation of both the model-generated and original stories. Their findings reveal that LLMs fall considerably short of \textit{experienced writers} in creating truly creative content. Tian et al. (\citeyear{Tian2024AreLL}) similarly finds that LLM-generated stories are \textit{non-diverse} and typically \textit{lack suspense} and \textit{tension}. However, these works largely focus on comparing models to award-winning professional writers while our work centers around comparing the creative story-writing abilities of models to average human writers. 

Past work closest to ours is the work of Orwig et al. (\citeyear{orwig2024language}) which also evaluates creative short story generation in both average humans and LLMs using the same five-story generation task. However, our work differs in several major aspects. First, Orwig et al. (\citeyear{orwig2024language}) compares a collection of humans to only a single model (either GPT-3 or GPT-4) where model story variation is achieved by varying temperature values. This setup makes an implicit assumption of treating the same model with a different decoding parameter as equal to an individual human. However, it remains unclear whether the same findings will hold if a population of different models is compared to a population of humans. Therefore, our study focuses on evaluating collections of both different humans and 60 different LLMs. Second, Orwig et al. (\citeyear{orwig2024language}) scores creativity with an overall rating and links content to different human memory processes. In our study, we conceptualize creativity as a multifaceted concept and characterize it by the dimensions of novelty, surprise and value \cite{Boden1991TheCM}. Finally, Orwig et al. (\citeyear{orwig2024language}) collects creativity ratings from non-expert raters and GPT-4 while our work considers ratings from both non-expert and expert human raters as well as three LLM judges. We further study where differences between these three types of judges in ratings come from, by predicting creativity ratings from automated metrics across multiple dimensions.

\begin{figure*}[h]
\begin{subfigure}[b]{0.5\textwidth}
    \centering
    \includegraphics[width=\textwidth]{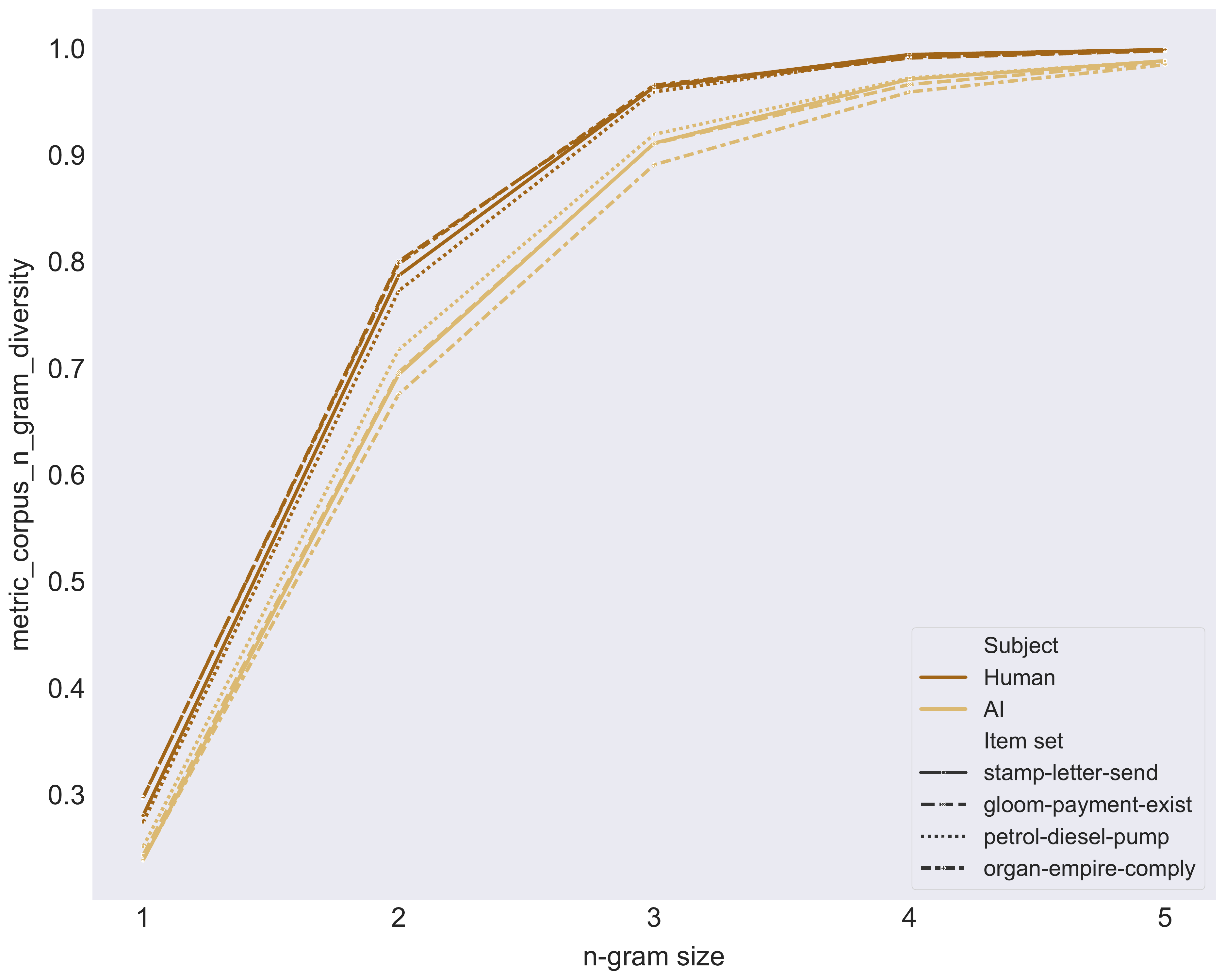}
    \caption{\textit{n}-gram diversity.}
    \label{fig:results-n-gram-divers}
\end{subfigure}
\begin{subfigure}[b]{0.5\textwidth}
    \centering
    \includegraphics[width=\textwidth]{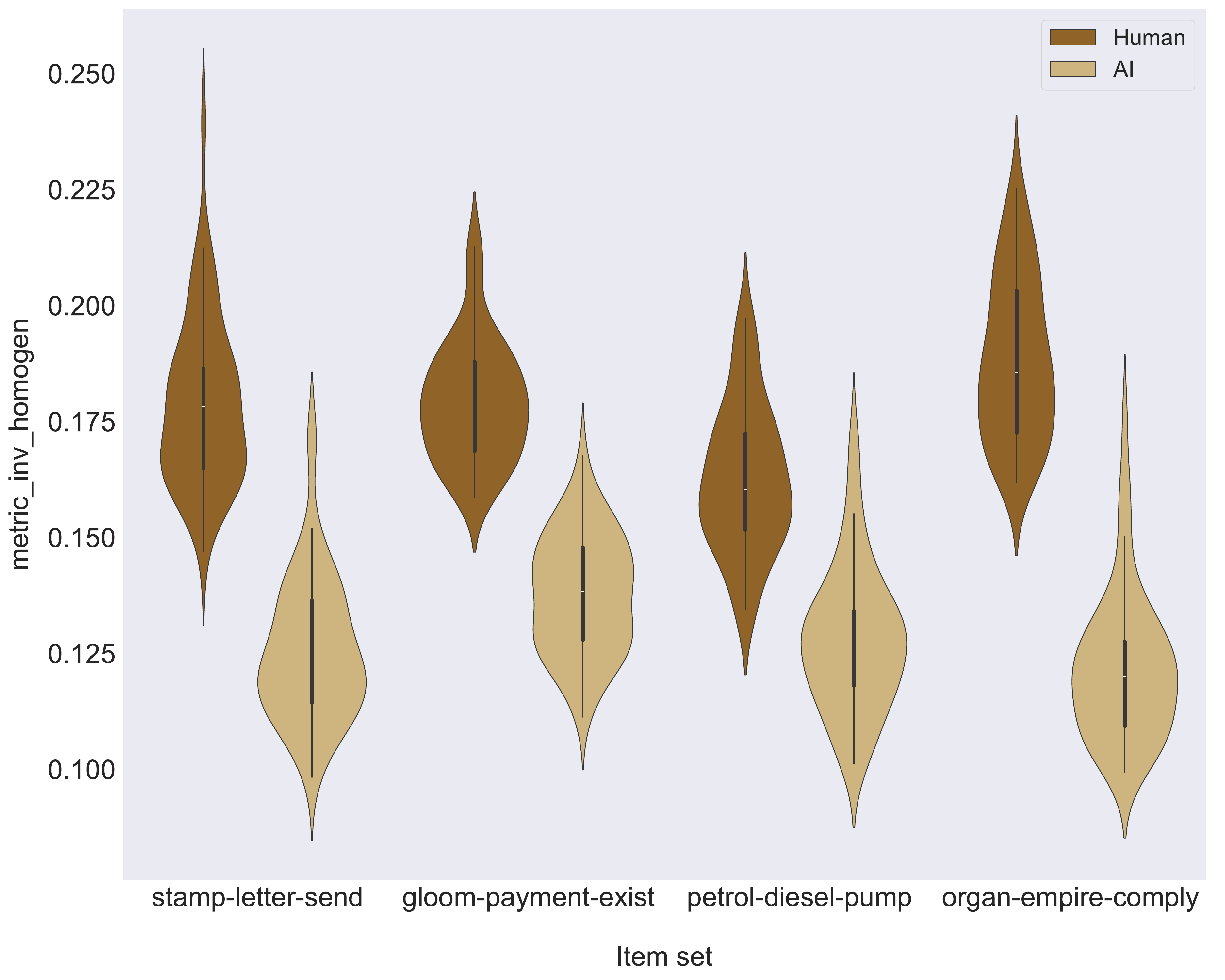}
    \caption{Inverse homogenization.}
    \label{fig:results-inv-homogen}
\end{subfigure}
\caption{Lexical and semantic diversity scores across all item sets measured by the \textit{n}-gram diversity and inverse homogenization metrics respectively.}
\label{fig:results-diversity}
\end{figure*}

%% file: 03_methodology.tex
\section{Methods}
\label{sec:methodology}

\subsection{Story Generation Data Collection}
We collected data from both humans and LLMs using a creative short story generation task based on three cue words e.g. \textit{stamp}, \textit{letter}, \textit{send} (also known as an item set). We chose this task because it is simple and often employed in psychology to assess human creativity in story generation \cite{prabhakaran2014thin,johnson2023divergent,orwig2024language}.
We use four sets of cue words from \cite{johnson2023divergent} where there is either a high semantic distance between words (\textit{gloom, payment, exist} and \textit{organ, empire, comply}) or a low semantic distance (\textit{stamp, letter, send} and \textit{petrol, diesel, pump}). Both humans and models were given the same instructions in English using the following prompt:

\begin{highlight}
\textbf{Instructions}\\
    You will be given three words (e.g., car, wheel, drive) and then asked to write a creative short story that contains these three words. The idea is that instead of writing a standard story such as "I went for a drive in my car with my hands on the steering wheel.", you come up with a novel and unique story that uses the required words in unconventional ways or settings.

    Write a creative short story using a maximum of five sentences. The story must include the following three words: \{items\}. However, the story should not be about \{boring\_storyline\}.
\end{highlight}

In the instructions above, \textit{items} refer to the cue words and \textit{boring\_storyline} corresponds to a typical or uncreative storyline that would first come to mind about those cue words. For example, a typical storyline for cue words \textit{stamp, letter, send} could be \textit{stamping a letter and sending it}. We include these hints in the instructions to increase the creativity of both human and LLM generated stories.

Human data were collected from $60$ participants ($43\%$ female, age: $M=38.8, SD=13.6$ years; fluent English speakers residing in the UK, with no language-related disorders and having completed secondary school education) on Prolific\footnote{https://www.prolific.com/}, a crowd-sourcing platform. Participants not adhering to instructions were removed, resulting in a total of $59$ participants.

For a fair comparison, model data were also collected from $60$ different models that are diverse in model size, training data and model architecture. The full list of models can be found in Models section. 
All models were prompted in zero-shot setting with a decoding setup that has been used in previous works to generate creative outputs  ($temperature=0.7, top\_p=0.95$) \cite{stevenson2022putting,nath2024characterising}.

In total, we collected $480$ stories ($60$ stories for humans and models each across $4$ item sets). To make sure all stories are meaningful and comparable in length, we performed some preprocessing to remove outlier stories that contain less than $3$ or more than $7$ sentences. This filtering step resulted in a total of $431$ stories for final evaluation. Final data statistics can be found in Appendix Table \ref{tab:final-data-stats}.

\subsection{Story Evaluation by Automated Metrics}
We evaluate both human and model stories using various automated metrics that correspond to different dimensions of creativity. These measures are either common methods relying on the basic linguistic structure of sentences (e.g. \textit{n}-grams, dependency trees) or metrics based on the notion of \textit{semantic distance} that has been shown as an effective automated metric to evaluate creativity \cite{Beaty2020AutomatingCA,Dunbar2009CreativityET,Harbison2014AutomatedSO,johnson2023divergent,prabhakaran2014thin,Karampiperis2014TowardsMF}. Semantic distance between two texts is typically computed based on the cosine similarity of embeddings of the texts. More specifically, we consider the following metrics:

\begin{figure}[h]
\centering
\includegraphics[width=\columnwidth]{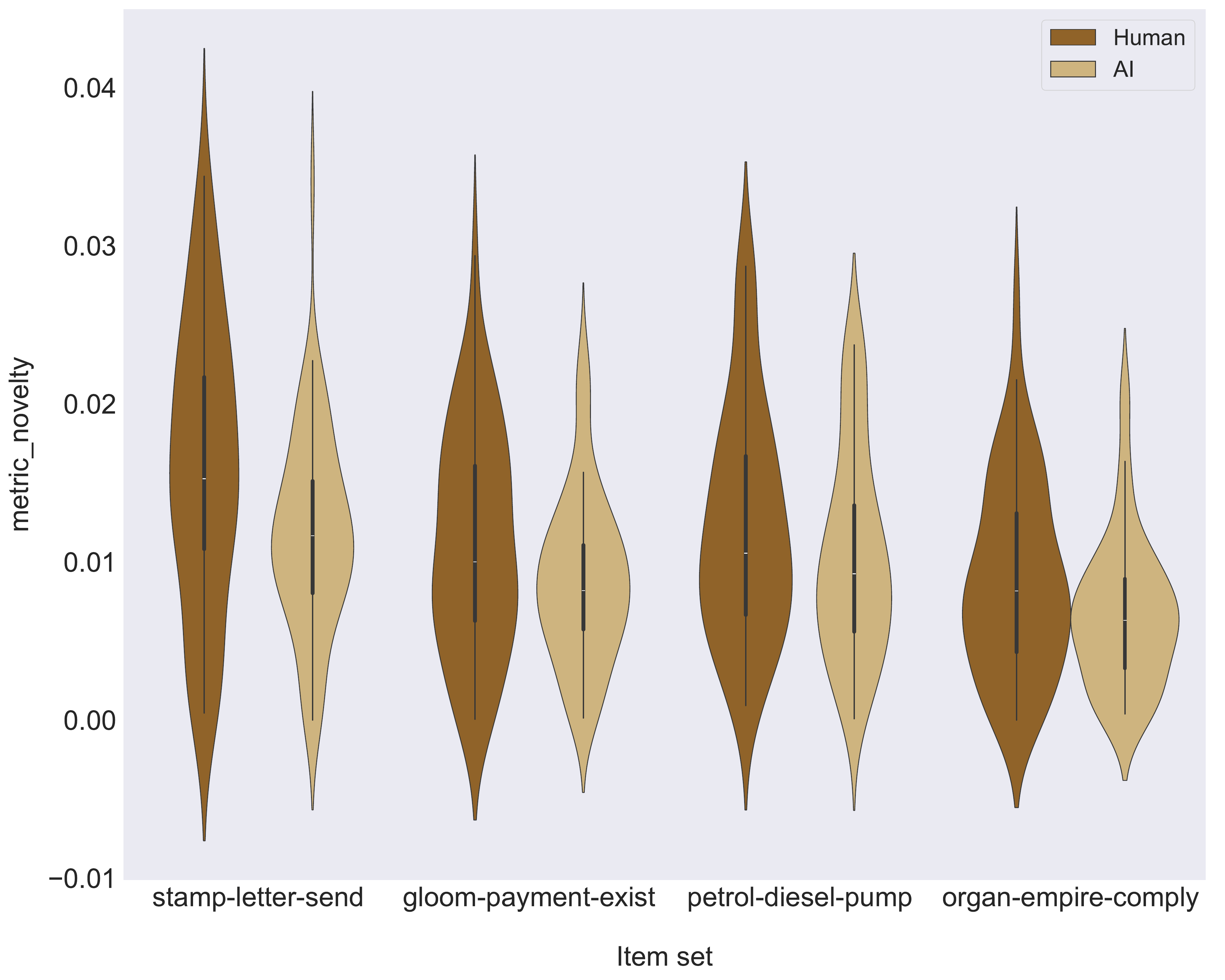}
\caption{Novelty scores across all item sets.}
\label{fig:results-novelty}
\end{figure}

\paragraph{Diversity} \label{method:diversity}
Creative stories are often characterized by diverse structures both at the lexical and semantic levels. To measure \textbf{lexical diversity}, we employ \textit{n}-gram diversity for values of \textit{n} from $1$ to $5$ where for a given \textit{n}, \textit{n}-gram diversity is defined as the ratio of the unique \textit{n}-grams to the total number of \textit{n}-grams in a story. To measure \textbf{semantic diversity}, we employ a diversity score similar to Padmakumar and He (\citeyear{padmakumar2023does}) which we call \textbf{inverse homogenization} score. It is defined as the average pairwise distance of a story to all other stories written on the same item set, i.e. $inv\_hom(s | t) = \frac{1}{|S_t|-1} \sum_{s' \in S_t \setminus s} semdis(s, s')$ where $S_t$ is a set of stories written on item set $t$ and $semdis$ corresponds to the semantic distance score. We use $1-cosine\_similarity$ as the semantic distance function and a sentence embedding model (\texttt{gte-large}) \cite{li2023generaltextembeddingsmultistage} to compute embeddings of stories.


\paragraph{Novelty} \label{method:novelty}
One of the major dimensions of creativity is the novelty aspect \cite{runco2012standard}. It is typically defined as the measure of how different an artifact is from other known artifacts in its class \cite{Maher2010EvaluatingCI}. To compute the novelty of a story, we employ the novelty metric from Karampiperis et al. (\citeyear{Karampiperis2014TowardsMF}) which defines it as the average semantic distance between the dominant terms (i.e. lemmatized content words) of the story, compared to the average semantic distance of the dominant terms in all stories. More formally, let $S_n$ be a given story, $S_G$ a corpus of all stories across all item-sets and $T_n$ and $T_G$ set of dominant terms respectively for $S_n$ and $S_G$. Then similar to Johnson et al. (\citeyear{johnson2023divergent}), we can define the average semantic distance between the dominant terms for $S_n$ as follows:

\begin{equation}
    D(S_n) = \frac{\sum_{i, j=1]}^{|T_n|}semdis(T_{ni}, T_{nj}), i\neq j}{|T_n|}
\end{equation}

and similarly for $S_G$ as follows:

\begin{equation}
    D(S_G) = \frac{\sum_{i, j=1]}^{|T_G|}semdis(T_i, T_j), i\neq j}{|T_G|}
\end{equation}

Then the novelty of the story $S_n$ can be defined as below (normalized to the $[0, 2]$ space):

\begin{equation}
    Nov(S_n) = 2|D(S_n) - D(S_G)|
\end{equation}

\paragraph{Surprise} \label{method:surprise}
Also known as unexpectedness, surprise has been shown to play an important role in characterizing a creative artifact \cite{Boden1991TheCM,grace2014expect}. It is typically defined as the artifact's degree of deviation from what is expected \cite{Maher2010EvaluatingCI}. In the context of a story, surprise can be induced as the story unfolds, i.e., the next sentence that deviates largely from the previous one can create an effect of surprise. Using this temporal dimension, Karampiperis et al. (\citeyear{Karampiperis2014TowardsMF}) defines the surprise of a story as the average semantic distances between the consecutive fragments (i.e. sentences) of each story, normalized in the $[0, 2]$ space. More formally, it could be defined as follows:

\begin{equation}
    Sur(S_n) = \frac{2}{|F|-1} \sum_{i=2}^{|F|}|D(F_i) - D(F_{i-1})|
\end{equation}

where $|F|$ refers to the number of fragments and $F_i$ is the $i$-th fragment. We employ this metric to compute a value of surprise for each generated story.

\begin{figure}[h]
    \centering
    \includegraphics[width=\columnwidth]{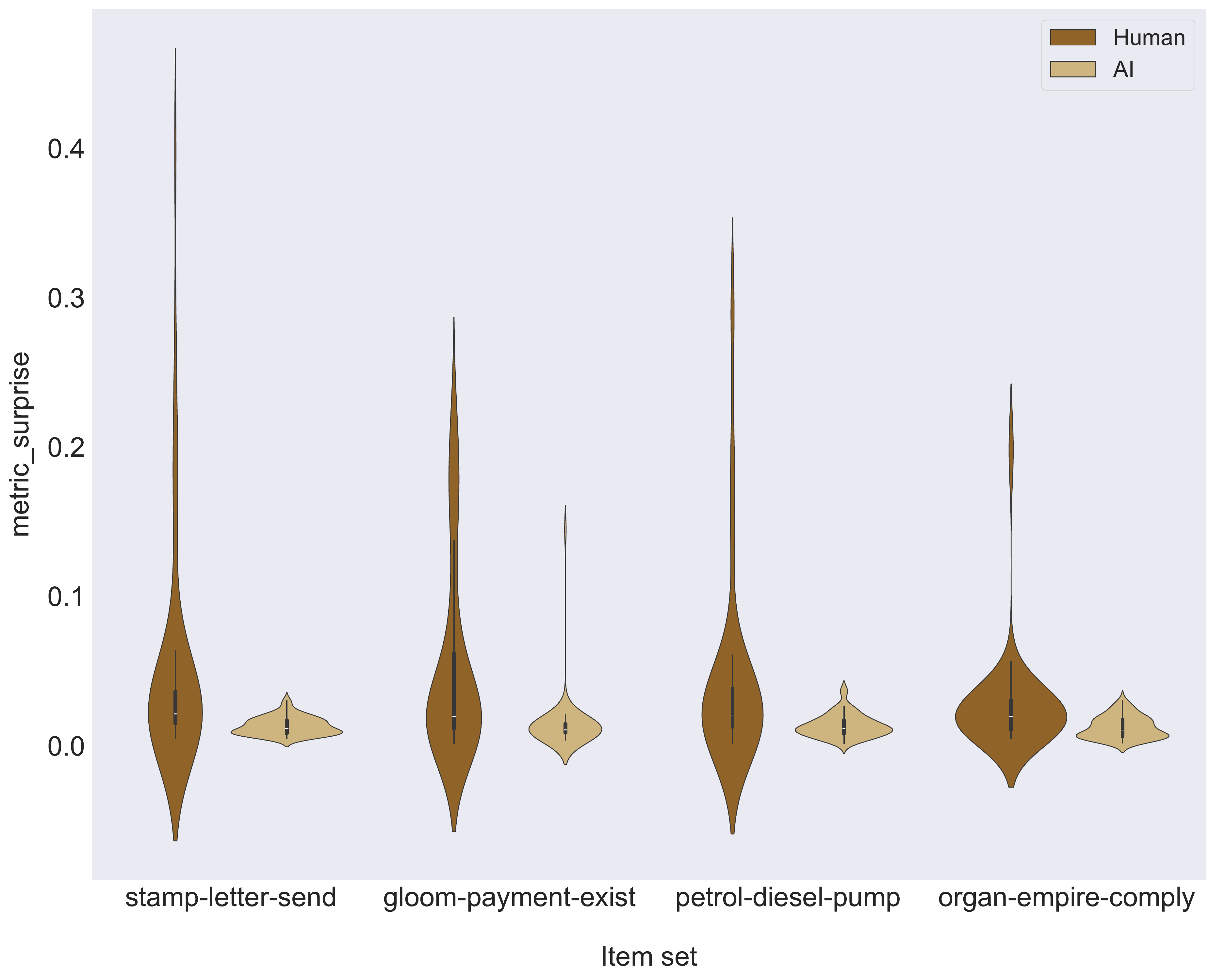}
    \caption{Surprise scores across all item sets.}
    \label{fig:results-surprise}
\end{figure}

\paragraph{Complexity} \label{method:complexity}
Finally, stylistic creativity can be injected into stories by making them linguistically complex. However, lexically and syntactically complex stories can often be unreadable or hard to follow for humans. To measure the level of complexity in generated stories, we employ \textbf{lexical} and \textbf{syntactic} complexity metrics. Lexical complexity metrics include \textit{number of unique words}, \textit{average word length}, \textit{average sentence length}, and \textit{readability}. For readability, we employ the Flesch reading ease score \cite{flesch1940}. Syntactic complexity metrics include \textit{part-of-speech tag ratios} (e.g. nouns, adjectives), \textit{average dependency path length}, and \textit{average constituency tree depth}. The average dependency path length is defined as the average of the lengths of dependency paths for each word in a sentence where a dependency path is a sequence of words that are connected with a dependency relation (e.g. \textit{subject of}). For example, in the sentence ``in the heart of an ancient library'', a dependency path corresponding to the word ``in'' would be \textit{in-heart-of-library} with a length of 4. The average constituency tree depth on the other hand is defined as the average of the lengths of the branches in a constituency tree of a sentence.

\begin{figure}[h]
    \centering
    \includegraphics[width=\columnwidth]{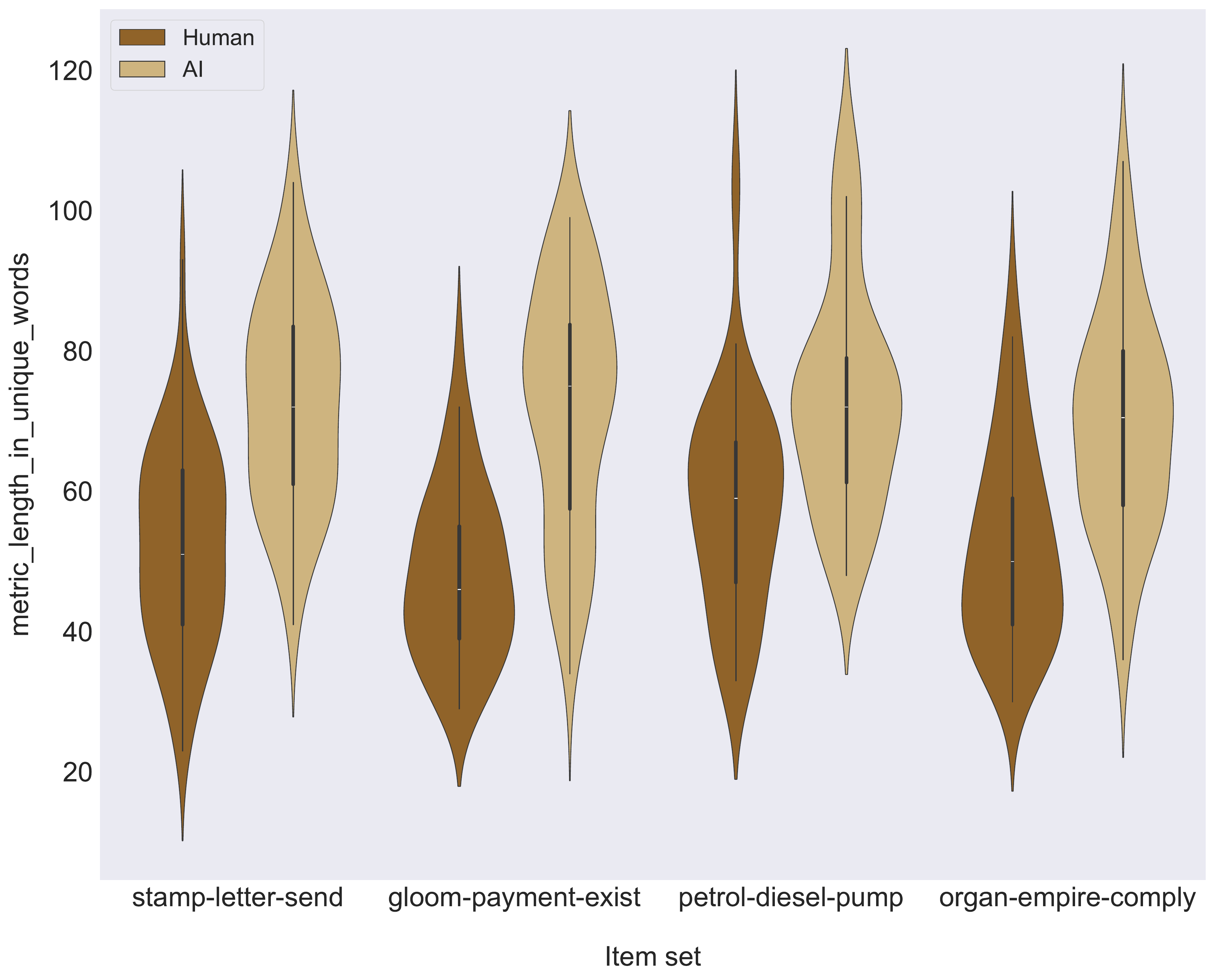}
    \caption{Lexical complexity scores across all item sets measured by story length in number of unique words.}
    \label{fig:results-story-length-in-words}
\end{figure}

\subsection{Story Evaluation by Judges}
Following the widely popular CAT method \cite{amabile1982social}, we evaluate both human and AI stories using non-expert and expert raters across several dimensions of creativity. In addition, we collect ratings from three LLMs that have been shown to be decent judges of many natural language processing tasks \cite{bavaresco2024llms}.

\paragraph{Human Expert Judges}
We had two trained research assistants—graduate psychology students with creative writing experience— score each of the $431$ valid stories on creativity, originality, surprise and effectiveness. For unbiased evaluation, these annotators were not involved in any stage of the study, were unfamiliar with the study design and were fully blinded to which stories stemmed from humans versus AI. Inspired by the original Turing test \cite{turing1950computing}, they were also asked to judge each story on whether it was created by human or an AI. The inter-rater reliability of their judgments ranged from good to excellent (ICC=$.62-.90$). Therefore, for each variable we compute composite scores by taking the means of the two human expert judges.

\begin{table*}[h]
    \centering
    \begin{tabular}{lclc}
         \toprule
         \multicolumn{2}{c}{{\textbf{Human}}} & \multicolumn{2}{c}{{\textbf{AI}}} \\
         \midrule
         \textbf{5-gram} & \textbf{Count} & \textbf{5-gram} & \textbf{Count} \\
         \midrule
         ``all she felt was gloom'' & 2 & ``in the heart of the'' & 33 \\
         ``hard to comply with the'' & 2 & ``in the heart of a'' & 20 \\
        ``went to the petrol station'' & 2 & ``the heart of a bustling'' & 13 \\
        ``a stamp from my collection'' & 2 & ``once upon a time in'' & 11 \\
        \bottomrule
    \end{tabular}
    \caption{Most frequent 5-grams in human and AI stories along with their repetition counts.}
    \label{tab:top-5-grams}
\end{table*}

\paragraph{Human Non-expert Judges}
We also conducted the same evaluation with an independent sample of $96$ non-expert judges recruited via Prolific ($49\%$ female, age: $M=39.8, SD=12.4$ years) The non-expert raters were -just like the recruited amateur writers- UK residents who spoke English fluently, had no language-related disorders and have completed secondary education. We collected 5 non-expert ratings for each story, the minimum required for reliable creativity judgments by non-experts \cite{long2022dissecting}. Due to time and cost constraints, we performed this evaluation study on a subset of the stories ($n=273$) that nonetheless cover all item sets. 

To ensure consistent and reliable evaluation, all annotators received detailed instructions on the definitions of creativity, novelty, surprise, and effectiveness. As often is the case in subjective creativity judgments, the variation in judgments across non-experts was much higher than between experts or LLMs \cite{long2022dissecting}, where the inter-rater reliability of their judgments ranged from fair to good (ICC=$.43-.71$). Therefore, we choose to use the median rating (i.e., most common rating given across all ratings of the story) rather than a mean to reduce the influence of outliers.

\paragraph{LLM Judges}
We also prompt three LLMs, i.e. Claude, Gemini and GPT-4, to rate each of the $431$ stories on the same five variables. LLM judges had excellent inter-rater reliability for all variables (ICC=$.86-.94$) except human vs AI judgments (ICC=$.43$, fair inter-rater agreement). Therefore, we compute the means of the LLM judges for creativity, originality, surprise and effectiveness. For human vs AI judgments  we take the median (i.e., most popular vote).

%% file: 04_results.tex
\section{Results}
\label{sec:results}

\subsection{Results of Story Evaluation by Automated Metrics}
In this section, we report and discuss the evaluation results using the automated metrics stratified by individual item sets. To measure the effect of the semantic distance within an item set on the creativity of the generated stories, we additionally report the automated metrics results stratified by the type of semantic distance (e.g. low vs. high).

\begin{figure}[h]
    \centering
    \includegraphics[width=\columnwidth]{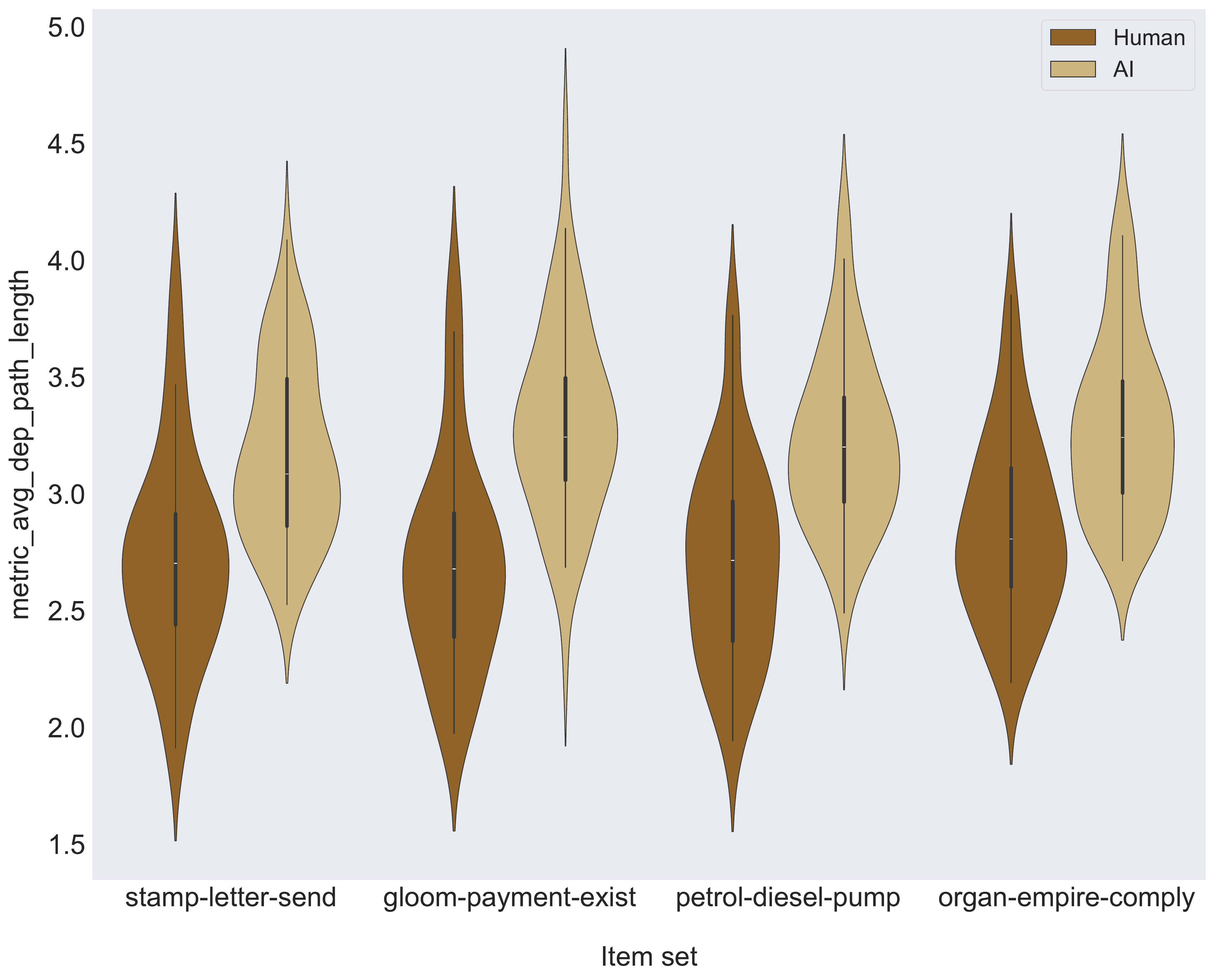}
    \caption{Syntactic complexity scores across all item sets measured by average dependency path length.}
    \label{fig:results-avg-dep-path-len}
\end{figure}

\paragraph{Diversity}
Figure \ref{fig:results-diversity} summarizes the results for lexical and semantic diversity metrics as measured by \textit{n}-gram diversity and inverse homogenization across all model and human groups and item sets. We see that humans consistently display a higher lexical and semantic diversity ($p < 0.0001$).

To get more insight into the type of \textit{n}-grams that are repeated across item sets, we report the frequency of most common \textit{5}-grams for both model and human stories in Table \ref{tab:top-5-grams}. We see that models tend to follow a story template by repeatedly using certain phrases while human stories exhibit no such behaviour.  

Moreover, inverse homogenization scores indicate that model stories tend to share the same themes while human stories are much more diverse in their content. To quantify the diversity of themes, similar to Nath et al. (\citeyear{nath2024characterising}) we perform agglomerative clustering using Ward variance minimization algorithm with a threshold of $0.6$ on the embeddings of all stories for a given item set which gives us a set of theme clusters. We find that human stories are characterized by significantly more themes than model stories (Figure \ref{fig:results-num-themes}). To further analyze what type of themes dominate model and human stories, for each group, we combine stories written on a given item set and ask GPT-4 to summarize them in a sentence. Results show that model stories tend to focus on themes of magical transformations or mysterious transactions while human stories speak of human nature, human interactions and responsibilities. Theme statistics and identified categories can be found in Appendix Figure \ref{fig:results-num-themes} and Table \ref{tab:themes} respectively.

We also analyze the effect of the item set semantic distance on the lexical and semantic diversity results. To do this, we stratify the \textit{n}-gram diversity and inverse homogenization scores across low and high semantic distance groups where we average \textit{n}-gram diversity scores over all \textit{n}-gram sizes (1-5). We find that while human stories written on high semantic distance items are more lexically and semantically diverse than those on low semantic distance items ($p < 0.001$), there is no significant difference between AI stories across different semantic distance categories. Results for this analysis can be found in Appendix Figure \ref{fig:results-diversity-semdis}.

\paragraph{Novelty}
Figure \ref{fig:results-novelty} summarizes the results of novelty metrics across all model and human groups and all item sets. We see that human stories are more novel than those of the models with varying levels of significance across item sets ($p < 0.01$, $p < 0.1$, $p < 0.05$ and $p < 0.05$). 

We also similarly analyze the effect of the item set semantic distance on the novelty scores. We observe that human and model stories corresponding to low semantic distance items exhibit more novelty ($p < 0.001$) than those of the high semantic distance items which also aligns with previous findings \cite{johnson2023divergent}. Results for this analysis can be found in Appendix Figure \ref{fig:results-novelty-semdis}.

\vspace{-5pt}
\paragraph{Surprise}
Figure \ref{fig:results-surprise} summarizes the results of surprise metrics across all model and human groups and all item sets. We see that human stories are more surprising ($p < 0.001$) than those of the models.

\begin{figure}[h]
\centering
\includegraphics[width=\columnwidth]{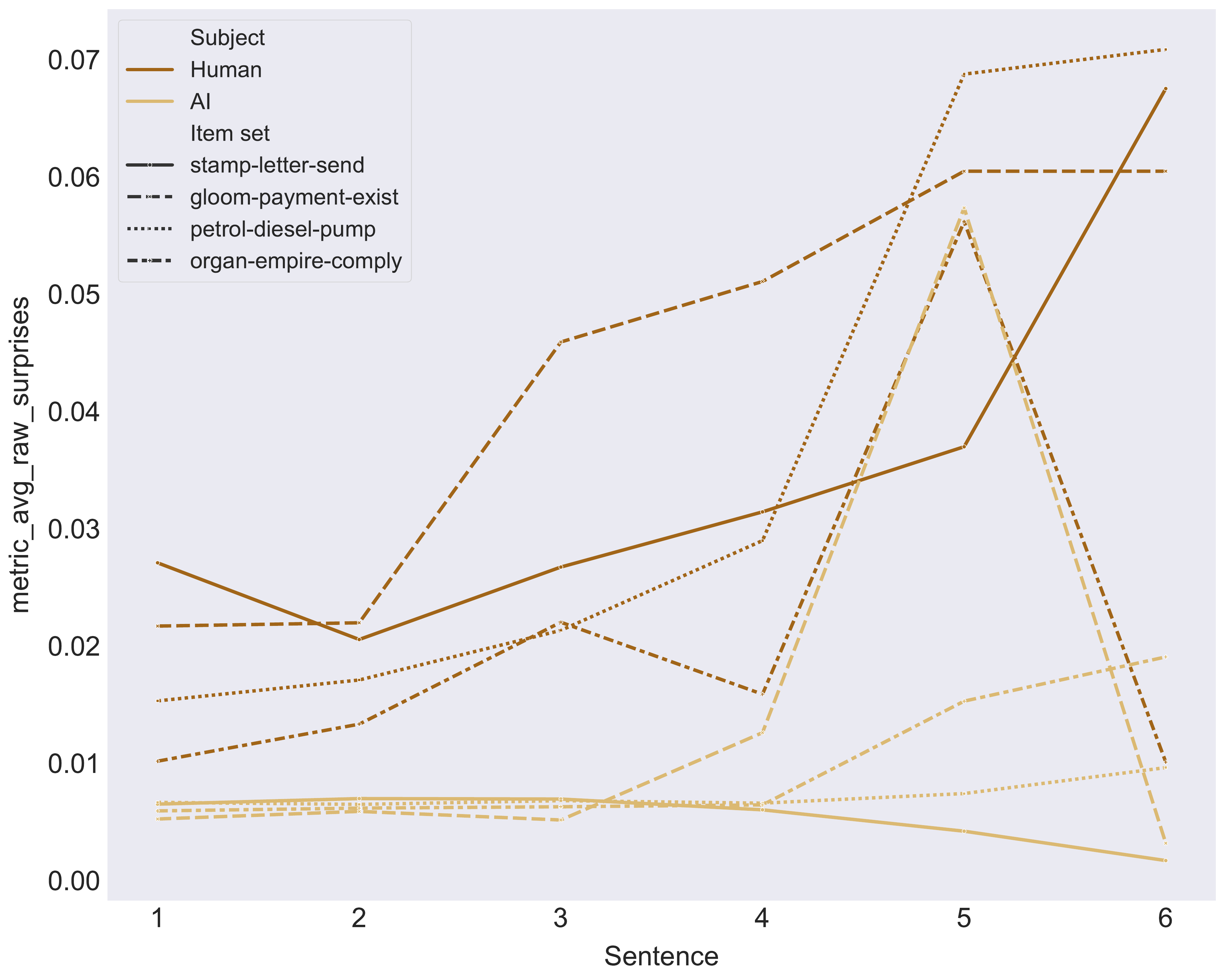}
\caption{Surprise profile scores across sentence positions averaged over all stories.}
\label{fig:results-surprise-profile}
\end{figure}

To analyze how the surprise changes as the story unfolds which we call the \textit{surprise profile} of a given model, we plot the averaged raw surprise scores across fragments (i.e. sentences) of the stories written on a given item set in Figure \ref{fig:results-surprise-profile}. We see that human stories exhibit greater surprise variation across sentences while model stories keep a largely monotonous profile. 

 We observe no significant difference between low and high semantic distance results for surprise. Results for this analysis are in Appendix Figure \ref{fig:results-surprise-semdis}.

\vspace{-5pt}
\paragraph{Complexity}
Figures \ref{fig:results-story-length-in-words} and \ref{fig:results-avg-dep-path-len} summarize some of the results for lexical and syntactic complexity metrics respectively across all model and human groups and item sets. What we see is that AI models consistently produce longer stories and their sentences are lexically and syntactically more complex as indicated by larger number of unique words and longer dependency paths per sentence ($p < 0.001$). Additionally, we find that models generally use more nouns and adjectives, while humans use more pronouns and adverbs ($p < 0.001$). To further analyze the type of pronouns used by humans and AI models, we perform an additional analysis on pronoun use and find that humans almost exclusively write their stories from the first or second person perspective, however, models prefer stories centered around third person. Overall, our findings show that models generally produce grammatically complex and potentially less readable stories. Results for pronoun analysis are in Appendix Figure \ref{fig:results-pronoun-use}. More complexity metric results are in Appendix Figures \ref{fig:results-pos-ratio} and \ref{fig:results-more-complexity}.

\begin{table*}[h]
  \centering
  \small
  \begin{tabular}{lcccccc}
    \toprule
     & \multicolumn{2}{c}{\textbf{Expert Judges}} & \multicolumn{2}{c}{\textbf{Non-expert Judges}} & \multicolumn{2}{c}{\textbf{LLM Judges}} \\
    \textbf{Creator:} & Human & AI & Human & AI & Human & AI \\
    \midrule
    creativity    & $3.58$ ($\pm0.78$) & $2.33$ ($\pm0.55$) & $2.45$ ($\pm0.77$) & $3.65$ ($\pm0.78$) & $2.41$ ($\pm0.70$) & $4.26$ ($\pm0.64$) \\
    originality   & $3.96$ ($\pm0.79$) & $2.70$ ($\pm0.73$) & $2.48$ ($\pm0.76$) & $3.47$ ($\pm0.75$) & $2.33$ ($\pm0.73$) & $4.22$ ($\pm0.72$) \\
    surprise      & $3.20$ ($\pm0.89$) & $1.68$ ($\pm0.61$) & $2.21$ ($\pm0.81$) & $2.96$ ($\pm0.78$) & $2.01$ ($\pm0.79$) & $3.74$ ($\pm0.77$) \\
    effectiveness & $3.87$ ($\pm0.71$) & $2.17$ ($\pm0.70$) & $2.70$ ($\pm0.73$) & $3.33$ ($\pm0.76$) & $2.86$ ($\pm0.58$) & $4.14$ ($\pm0.52$) \\
    \bottomrule
  \end{tabular}
  \caption{Means ($\pm$SDs) of aggregated ratings by expert, non-expert and LLM judges.}
  \label{tbl:crearatingpredictors}
\end{table*}

When we analyze the effect of the item set semantic distance on the complexity scores, we observe a significant difference only with respect to lexical complexity scores for human stories where low semantic distance item sets result in more lexically complex stories than high semantic distance item sets ($p < 0.01$). Results for this analysis are in Appendix Figure \ref{fig:results-semdis-complexity}.

\subsection{Results of Story Evaluation by Judges}
The descriptive statistics of the mean ratings per variable can be found in Table \ref{tbl:crearatingpredictors}. For each group of judges the ratings for creativity, originality, surprise and effectiveness are all highly correlated (experts: $.83<r<.89$; non-experts: $.64<r<.74$; LLMs: $.92<r<.99$). Therefore, we focus solely on the creativity ratings and human vs AI judgments for further analyses.

\paragraph{Who's more creative according to our judges?}
Experts rate human generated stories $1.25$ points higher (on a scale of 5) than those generated by LLMs ($t=-19.34, p<.001$). In contrast, both non-experts and LLM judges give AI-generated stories higher scores, where non-experts rate AI stories with $1.19$ more points ($t=12.76, p<.001$) and LLM judges rate AI stories with $1.85$ more points ($t=28.75, p<.001$).  

\paragraph{Turing Test: Could LLMs fool our judges and pass for humans?}
Experts predict the author (human or AI) correctly $94\%$ of the time, outperforming both non-expert and LLM judges. Non-experts make the correct prediction $81\%$ of the time and LLM judges predict human vs. AI with $71\%$ accuracy. Additionally, to gain insight into what factors drive human judgment on whether a story is written by AI or a human, we ask the non-experts to explain their strategy to predict the author of a story. We summarize their comments into several themes of qualities that were attributed the most respectively to human and AI stories. We find that qualities identified by non-experts coincide with our findings from the automated metric analysis that AI models tend to produce linguistically more complex and verbose yet less creative stories than humans (Table \ref{tab:non-expert-qualities}).

\subsection{Which automated metrics predict creativity evaluation by the three groups of judges?}
We create regression models to predict creativity ratings generated by the three different judges. We choose to use a simple explainable model with one predictor for each category of automated metrics. For \textit{lexical diversity} we use \texttt{mean n-gram diversity} of a story and for \textit{semantic diversity} we use \texttt{inverse homogenization} (see Methods section for definitions). For \textit{novelty} and \textit{surprise} we use the metrics as described in the Methods section. For syntactic and lexical complexity, since all metrics are highly correlated, we choose a single metric that correlates strongest with creativity ratings to represent each construct. For \textit{syntactic complexity} we select \texttt{average constituency tree depth} and for \textit{lexical complexity} we select \texttt{number of unique words}. 

Using these metrics we run the following regression analysis to predict creativity ratings for stories by the three groups of judges: $creativity \sim semantic\_diversity + lexical\_diversity + novelty + surprise + syntactic\_complexity + lexical\_complexity$. 

As can be seen in Table \ref{tbl:crearatingpredictors2}, expert creativity ratings are best predicted by higher semantic diversity, surprise and lexical diversity scores. For non-experts, creativity ratings increase with higher lexical complexity (i.e., number of unique words). Surprisingly, non-expert creativity ratings decrease with higher semantic diversity, novelty and surprise scores. For LLM judges, we see the same pattern of predictors as for non-experts, where the best positive predictor of creativity ratings is lexical complexity and that semantic diversity, novelty and surprise are each negatively related to creativity ratings.

%% file: 06_discussion.tex
\section{Discussion}

\begin{table*}[t]
  \centering
  \begin{tabular}{lrrrrrrrrr}
    \toprule
     & \multicolumn{3}{c}{{\textbf{Expert Judges}}} & \multicolumn{3}{c}{{\textbf{Non-expert Judges}}} & \multicolumn{3}{c}{{\textbf{LLM Judges}}} \\
    \midrule
    \textbf{predictor} & \textbf{$\beta (SE)$} & \textbf{$t$} & \textbf{$p$} & \textbf{$\beta (SE)$} & \textbf{$t$} & \textbf{$p$} & \textbf{$\beta (SE)$} & \textbf{$t$} & \textbf{$p$}  \\
    \midrule
    $semantic\_diversity$   &  .56 (.04) & 14.65 & *** &   -.32 (.05) & -6.31 & *** & -.58 (.04) & -14.99 & *** \\ 
    $lexical\_diversity$    &  .09 (.04) &  2.32 & *   &    .01 (.05) &  0.26 &     &  .05 (.04) &  1.21  &     \\
    $novelty$               & -.03 (.03) & -0.80 &     &   -.20 (.05) & -4.48 & *** & -.21 (.04) & -5.87  & *** \\
    $surprise$              &  .09 (.04) &  2.36 & *   &   -.08 (.05) & -1.57 &     & -.08 (.04) & -2.10  & *   \\ 
    $syntactic\_complexity$ & -.04 (.04) & -1.01 &     & -.0006 (.06) & -0.10 &     & -.06 (.04) & -1.35  &     \\
    $lexical\_complexity$   &  .03 (.05) &  0.58 &     &    .37 (.06) &  6.06 & *** &  .42 (.05) &  8.78  & *** \\
    \bottomrule
  \end{tabular}
  \caption{Regression coefficients (SE) and t-test results for predictions of creativity ratings for each group of judges. 
  Where \textit{p}-value significance is represented as follows: ‘***’$<$.001 , ‘**’$<$.01 , ‘*’$<$.05 , and ‘ ’$>=$.05.}
  \label{tbl:crearatingpredictors2}
\end{table*}

In this work, we study and compare the creative short story generation abilities of humans and LLMs using a five-sentence short story generation task based on cue words. We use both automated metrics and judgments of non-expert and expert humans as well as LLMs to evaluate the creativity of the stories across several dimensions such as novelty, surprise, diversity, and complexity. For the complexity measures, we leverage common metrics relying on the linguistic structures of the stories at both lexical and syntactic levels. Then we analyze the results across all item sets and study the similarities and differences between the evaluations of different judges. 

Our analysis using the automated metrics shows that LLMs produce linguistically and stylistically more complex stories than humans as indicated by higher lexical and syntactic complexity results. However, human stories consistently exhibit higher novelty, surprise, and lexical and semantic diversity while being linguistically much less complex and easier to read. Our findings are in line with some previous work comparing LLM creativity to humans in story generation \cite{Chakrabarty2023ArtOA,Tian2024AreLL,Marco2024PronVP,Marco2024SmallLM} and creative problem-solving \cite{Tian2023MacGyverAL}. On the other hand, some past works have found no significant difference between overall LLM and human creativity in short story generation when comparing a population of humans to GPT-3 and GPT-4 when judged by non-experts and GPT-4 \cite{orwig2024language}. However, our fine-grained analysis considering multiple dimensions of creativity and a population of 60 different LLMs evaluated by automated metrics and experts reveals significant gaps between human and LLM stories across all major dimensions of creativity in favor of humans. 

Particularly, we find that our automated metric results highly correlate with expert judgments, while LLM and non-expert judgments tend to rate LLM stories as more creative than human stories. Our further analysis shows that this discrepancy stems from the underlying factors driving the different judgments. More specifically, expert judgments are driven by the diversity and surprise aspects of the stories which are essential to creativity while non-expert and LLM judgments highly correlate with sheer lexical complexity such as the number of unique words, which does not necessarily imply semantic complexity. In fact, our parts-of-speech analysis shows that LLMs tend to overuse rare adjectives and complex syntactical structures. Moreover, past works generally find expert judgments as more reliable evaluators of creativity than those of non-experts \cite{kaufman2012beyond,hu2023comparing,long2022dissecting,ceh2022assessing,veale2015game,lamb2015human,gervas2019exploring}. Similarly, LLMs-as-judges have been shown to be unreliable \cite{Chakrabarty2023ArtOA,Chhun2024DoLM} and biased towards their own generations \cite{wataoka2024self,panickssery2025llm}. 

\begin{figure}[h]
\centering
\includegraphics[width=\columnwidth]{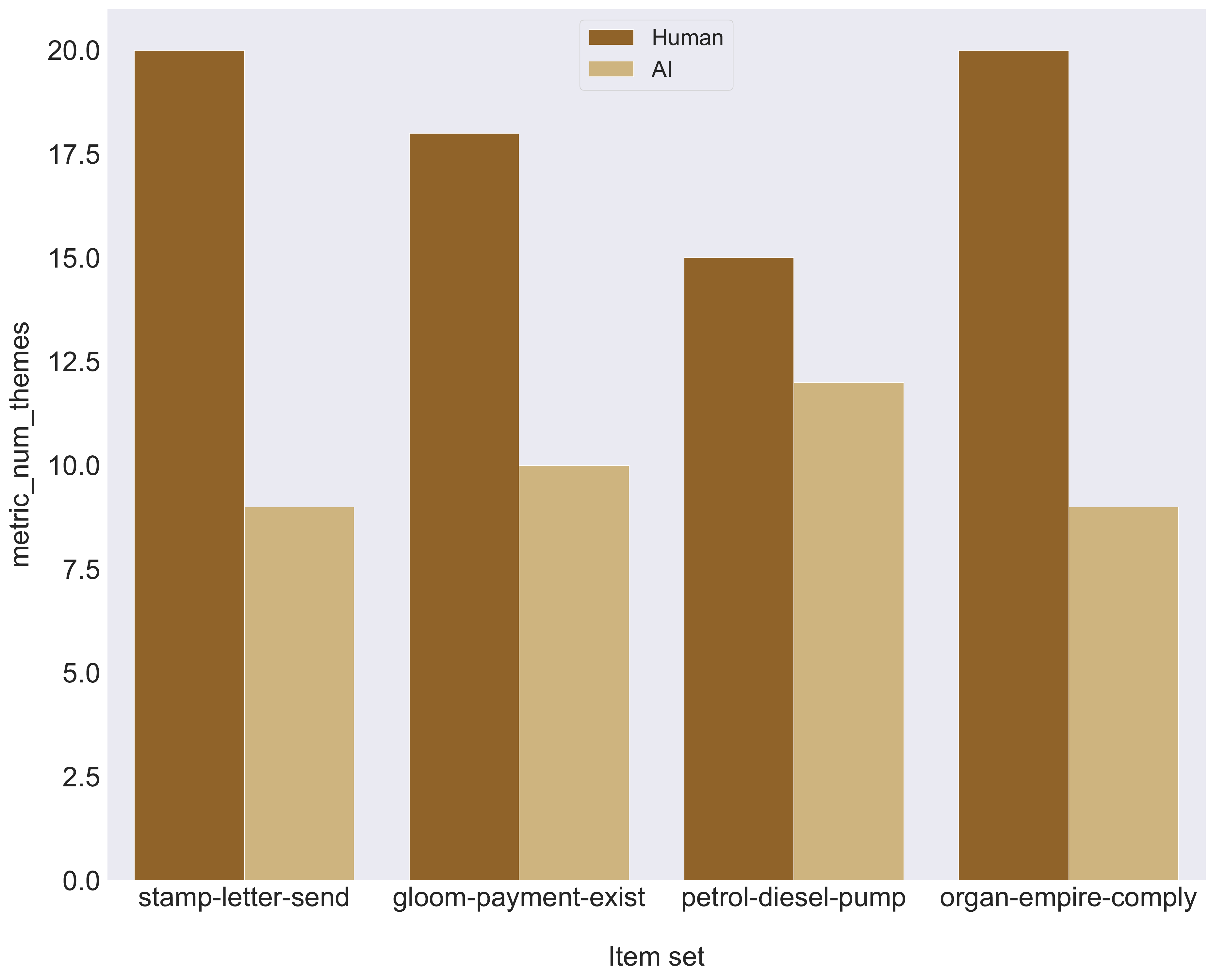}
\caption{Number of themes for human and AI stories.}
\label{fig:results-num-themes}
\end{figure}

Our work has several implications. The complexity vs. creativity gap shows that humans and LLMs have different interpretations of what it means to be creative for stories. While humans prefer telling a simple story from their perspective that is nonetheless surprising and original, LLMs, however, represent creativity with lexically and syntactically overloaded sentences narrated from the third person perspective and that typically focus on a few repetitive themes. Additionally, the fact that non-experts and LLMs tend to evaluate AI stories as more creative could mean that complexity creates the illusion of being more creative to the untrained eye. This behaviour can be due to several factors involved in training LLMs such as the training data, pre-training and post-training optimizations. For example, aligning LLMs with human feedback to be more helpful has been attributed to result in strong verbosity bias \cite{saito2023verbosity} and diversity reduction \cite{padmakumar2023does} in creative tasks. 
Our findings call for a more comprehensive evaluation of creativity and can inform future work on designing methods to improve the creativity of LLMs \cite{ismayilzada2024creativity}. Potential directions can include developing new prompt engineering \cite{Mehrotra2024EnhancingCI,Nair2024CreativePS,SummersStay2023BrainstormTS,Tian2023MacGyverAL} or optimization techniques \cite{Broad2021ActiveDW,Bunescu2019LearningTS,elgammal2017can} or steering internal mechanisms of LLMs using mechanistic interpretability \cite{Bereska2024MechanisticIF}.

%% file: 08_limitations.tex
\section*{Limitations}
\label{sec:limitations}
While our study provides a comprehensive analysis of a wide range of language models using fine-grained creativity metrics, it does have a few limitations. First, although all models were prompted using a decoding setup that has been used in previous work to favor creative output and idea diversity \cite{stevenson2022putting,nath2024characterising}, we did not explore alternative decoding or prompting strategies due to the high cost of open-ended evaluation across many models. 
Second, the novelty metric we employ is reference-based, meaning its outcomes depend heavily on the chosen reference corpus. In our case, this was the set of human- and AI-written stories for each item set. Using reference-based scoring is a shortcoming in nearly all creativity research that uses uniqueness to define novelty \cite{silvia2008assessing}. However, the construct validity of frequency-based uniqueness assessments is also high and their generalizability to stories from other populations increases with sample size \cite{lee2008commentary}. Therefore, future studies could validate our findings by including more stories in reference-based metrics.
Third, we used the minimum recommended number of expert and non-expert raters per story, i.e., two for (quasi-)experts —as these are generally highly reliable as in our study— and five non-experts —as these are generally less reliable— \cite{long2022dissecting}. Having more raters could have improved the reliability of our findings. Furthermore, including professional creative writers as experts could perhaps provide further improve reliability (although research suggests that quasi-experts are as reliable as experts \cite{long2022dissecting}, sometimes more so \cite{tan2015judgements}).
Lastly, although our creativity metrics are effective for evaluating story generation, they cannot fully capture the broader cultural and social dimensions of creativity or the depth of truly original and imaginative language use.

%% file: 09_ethics.tex
\section*{Ethics}
\label{sec:ethics}
All authors declare no conflicts of interest. No artificial intelligence assisted technologies were used in this research or the creation of this article. This research received approval from a local ethics board on September 11, 2024 (ID: FMG-10319).
All study materials are publicly available\footnote{https://github.com/mismayil/creative-story-gen}.

%% file: 10_acknowledgements.tex
\section*{Acknowledgements}
\label{sec:acknowledgements}
This publication is part of the project C\_LING (grant 205121\_207437, Swiss National Science Foundation) awarded to Lonneke van der Plas. 

%% file: 11_appendix.tex
\begin{table*}[t]
    \centering
    \begin{tabular}{p{2cm} p{7cm} p{7cm}}
         \toprule
         \textbf{Theme} & \textbf{Human Story Qualities} & \textbf{AI Story Qualities} \\
         \midrule
         Emotional depth & \apricothl{``emotional''}, \apricothl{``relatable''}, \apricothl{``personal''}, \apricothl{``first-person''}, \apricothl{``evoking feelings''}, \apricothl{``empathy''} & \apricothl{``formulaic format''}, \apricothl{``more abstract''}, \apricothl{``bad flow''}, \apricothl{``no feeling''}, \apricothl{``no depth''} \\
         \midrule
         Verbosity & \skybluehl{``simple''}, \skybluehl{``shorter''}, \skybluehl{``to the point''}, \skybluehl{``casual''}, \skybluehl{``pragmatic''}, \redviolethl{``less adjectives''} &  \skybluehl{``verbose''}, \skybluehl{``wordy''}, \skybluehl{``elaborate''}, \skybluehl{``descriptive''}, \melonhl{``repeating phrases''} (``once upon a time'', ``'in a world where '), \redviolethl{``lots of pointless adjectives''} \\
         \midrule
         Enjoyability & \orchidhl{``more depth''}, \orchidhl{``more sense''}, \orchidhl{``rythmic''}, \orchidhl{``easier to follow''}, \orchidhl{``more enjoyable''} & \orchidhl{``gibberish''},  \orchidhl{``hard to read''}, \orchidhl{``nonsense''}, \orchidhl{``convoluted''} \\
         \midrule
         Plausibility & \redviolethl{``everyday life''}, \redviolethl{``mundane``}, \redviolethl{``spelling and grammar errors''},  & \redviolethl{``fantastic''}, \redviolethl{``unrealistic''}, \redviolethl{``far-fetched''} \\
        \bottomrule
    \end{tabular}
    \caption{Most attributed qualities to human and AI stories by non-experts grouped by shared themes.}
    \label{tab:non-expert-qualities}
\end{table*}

\begin{table*}[h]
    \centering
    \begin{tabular}{p{4cm} p{6cm} p{6cm}}
         \toprule
         \textbf{Item set} & \textbf{Human Theme} & \textbf{AI Theme} \\
         \midrule
         stamp, letter, send & ``stamps as a symbol of connection, \redhl{serendipity}, and personal significance exploring themes of nostalgia, discovery, and \redhl{human interaction}''  & ``\greenhl{transformative} power of communication, where \greenhl{magical} stamps, letters, or trees serve as conduits for emotions, dreams, and connections, bridging distances and revealing truths that unite hearts, \greenhl{inspire} change, or unlock hidden potential.'' \\
         \midrule
         petrol, diesel, pump & ``\redhl{missteps}, chance encounters, and unforeseen consequences highlighting the quirks of \redhl{human nature}'' & ``\greenhl{transformative} and \greenhl{symbolic} nature of fuel—whether petrol, diesel, or emotions—serving as a metaphor for change, connection, and renewal in a world where innovation, \greenhl{magic}, and human spirit ignite hope and forge new beginnings'' \\
         \midrule
         gloom, payment, exist & ``\redhl{existential questioning}, the monotony of routine, and the weight of obligations, with "payments" and "gloom" serving as metaphors for the emotional toll of \redhl{life’s responsibilities}'' & ``\greenhl{negotiating existence} through transactions or sacrifices, often set against a backdrop of gloom that symbolizes darkness, \greenhl{mystery}, or transformation.'' \\
         \midrule
         organ, empire, comply & ``\redhl{power, control}, and the juxtaposition of grandeur with conformity, using the organ as a central symbol—representing both the majesty of empires and the \redhl{weight of their demands}.'' & ``music as a powerful force that shapes empires, demanding compliance or \greenhl{inspiring rebellion}, with the organ serving as a metaphor for unity, defiance, and the delicate balance between harmony and individuality.'' \\
        \bottomrule
    \end{tabular}
    \caption{Dominating themes for human and AI stories across all item sets.}
    \label{tab:themes}
\end{table*}

\begin{table*}[]
    \centering
    \begin{tabular}{ccccc}
    \toprule
    \textbf{Model name} & \textbf{Model version} & \textbf{Model size} & \textbf{Model author} \\
    \midrule
    GPT-3.5 & \texttt{gpt-3.5-turbo} & 175B & OpenAI \\
    GPT-4 & \texttt{gpt-4} & Unknown & OpenAI \\
    GPT-4o & \texttt{gpt-4o} & Unknown & OpenAI \\
    \midrule
    Claude 3 Opus & \texttt{claude-3-opus-20240229} & Unknown & Anthropic \\
    Claude 3.5 Sonnet & \texttt{claude-3-5-sonnet-20240620} & Unknown & Anthropic \\
    Claude 3.5 Haiku & \texttt{claude-3-5-haiku-20241022} & Unknown & Anthropic \\
    \midrule
    Gemini 1.5 Flash & \texttt{gemini-1.5-flash} & Unknown & Google \\
    Gemini 1.5 Pro & \texttt{gemini-1.5-pro} & Unknown & Google \\
    Gemma 2 9B & \texttt{gemma-2-9b} & 9B & Google \\
    Gemma 2 27B & \texttt{gemma-2-27b} & 27B & Google \\
    \midrule
    Llama 3.2 1B & \texttt{llama-3.2-1b-instruct} & 1B & Meta \\
    Llama 3.2 3B & \texttt{llama-3.2-3b-instruct} & 3B & Meta \\
    Llama 3.1 8B & \texttt{llama-3.1-8b-instruct} & 8B & Meta \\
    Llama 3.1 70B & \texttt{llama-3.1-70b-instruct} & 70B & Meta \\
    Llama 3.1 405B & \texttt{llama-3.1-405b-instruct} & 405B & Meta \\
    \midrule
    Grok 2 & \texttt{grok-beta} & 314B & xAI \\
    \midrule
    MPT 7B & \texttt{mpt-7b-8k-chat} & 7B & Databricks \\
    MPT 30B & \texttt{mpt-30b-chat} & 30B & Databricks \\
    DBRX & \texttt{dbrx-instruct} & 132B & Databricks \\
    \midrule
    DeepSeek LLM 7B & \texttt{deepseek-llm-7b-chat} & 7B & DeepSeek AI \\
    DeepSeek LLM 67B & \texttt{deepseek-llm-67b-chat} & 67B & DeepSeek AI \\
    \midrule
    Ministral 3B & \texttt{ministral-3b-instruct} & 3B & Mistral AI \\
    Ministral 8B & \texttt{ministral-8b-instruct} & 8B & Mistral AI \\
    Mistral 7B & \texttt{mistral-7b-instruct-v0.3} & 7B & Mistral AI \\
    Mistral Nemo 12B & \texttt{mistral-nemo-12b-instruct} & 12B & Mistral AI \\
    Mistral Small & \texttt{mistral-small-instruct-2409} & 22B & Mistral AI \\
    Mistral Large & \texttt{mistral-large-instruct-2407} & 123B & Mistral AI \\
    Mixtral 8x7B & \texttt{mixtral-8x7b-instruct} & 13B$^*$ & Mistral AI \\
    Mixtral 8x22B & \texttt{mixtral-8x22b-instruct} & 39B$^*$ & Mistral AI \\
    \midrule
    Nous Hermes 2 & \texttt{nous-hermes-2-mixtral-8x7b-dpo} & 13B$^*$ & Nous Research \\
    \midrule
    Qwen2.5 7B & \texttt{qwen-2.5-7b-instruct} & 7B & Qwen Team \\
    Qwen2.5 72B & \texttt{qwen-2.5-72b-instruct} & 72B & Qwen Team \\
    Qwen2.5 Coder 32B & \texttt{qwen-2.5-coder-32b-instruct} & 32B & Qwen Team \\
    \midrule
    Reka Edge & \texttt{reka-edge} & 7B & Reka AI \\
    Reka Flash & \texttt{reka-flash} & 21B & Reka AI \\
    Reka Core & \texttt{reka-core} & 67B & Reka AI \\
    \midrule
    Solar 10.7B & \texttt{solar-10.7b-instruct-v1.0} & 10.7B & Upstage AI \\
    \midrule
    GLM-4 & \texttt{glm-4-0520} & 130B & Zhipu AI \\
    \midrule
    Jamba-1.5-Mini & \texttt{jamba-1.5-mini} & 12B$^*$ & AI21 Labs \\
    Jamba-1.5-Large & \texttt{jamba-1.5-large} & 94B$^*$ & AI21 Labs \\
    \midrule
    Phi-3-Mini & \texttt{Phi-3-mini-4k-instruct} & 3.8B & Microsoft \\
    Phi-3-Small & \texttt{Phi-3-small-8k-instruct} & 7B & Microsoft \\
    Phi-3-Medium & \texttt{Phi-3-medium-4k-instruct} & 14B & Microsoft \\
    Phi-3.5-MoE & \texttt{Phi-3.5-MoE-instruct} & 6.6B$^*$ & Microsoft \\
    \midrule
    Aya Expanse 8B & \texttt{c4ai-aya-expanse-8b} & 8B & Cohere \\
    Aya Expanse 32B & \texttt{c4ai-aya-expanse-32b} & 32B & Cohere \\
    Command R+ & \texttt{command-r-plus} & 104B & Cohere \\
    \bottomrule
    \end{tabular}
    \caption{List of AI models evaluated in our study. Model size corresponds to number of parameters in billions. $^*$Active parameter size for Mixture-of-Experts (MoE) models. The rest of the models are listed in Table \ref{tab:models2}.}
    \label{tab:models}
\end{table*}

\begin{table*}[]
    \centering
    \begin{tabular}{ccccc}
    \toprule
    \textbf{Model name} & \textbf{Model version} & \textbf{Model size} & \textbf{Model author} \\
    \midrule
    Nemotron Mini & \texttt{nemotron-mini-4b-instruct} & 4B & Nvidia \\
    Nemotron-4 & \texttt{nemotron-4-340b-instruct} & 340B & Nvidia \\
    \midrule
    Yi-1.5 9B & \texttt{yi-1.5-9b-chat} & 9B & 01-AI \\
    Yi-1.5 34B & \texttt{yi-1.5-34b-chat} & 34B & 01-AI \\
    \midrule
    Baichuan 2 7B & \texttt{baichuan2-7b-chat} & 7B & Baichuan AI \\
    Baichuan 2 13B & \texttt{baichuan2-13b-chat} & 13B & Baichuan AI \\
    \midrule
    Zamba 2 7B & \texttt{zamba2-7b-instruct} & 7B & Zyphra \\
    \midrule
    Granite 3.0 2B & \texttt{granite-3.0-2b-instruct} & 2B & IBM \\
    Granite 3.0 8B & \texttt{granite-3.0-8b-instruct} & 8B & IBM \\
    \midrule
    StableLM Zephyr 3B & \texttt{stablelm-zephyr-3b} & 3B & Stability AI \\
    StableLM 2 12B & \texttt{stablelm-2-12b-chat} & 12B & Stability AI \\
    \midrule
    OLMo 2 7B & \texttt{olmo-2-7b} & 7B & Allen AI \\
    OLMo 2 13B & \texttt{olmo-2-13b} & 13B & Allen AI \\
    \midrule
    LFM 40B & \texttt{lfm-40b} & 40B & Liquid AI \\
    \bottomrule
    \end{tabular}
    \caption{List of AI models (continued) evaluated in our study. Model size corresponds to number of parameters in billions. $^*$Active parameter size for Mixture-of-Experts (MoE) models.}
    \label{tab:models2}
\end{table*}

\begin{table*}[h]
    \centering
    \begin{tabular}{p{4cm} p{6cm} p{6cm}}
         \toprule
         \textbf{Item set} & \textbf{Human Story} & \textbf{AI Story} \\
         \midrule
         stamp, letter, send & ``Send for the detective'' screamed the woman staring at the body on the ground. Detective Monroe soon arrived and could clearly see the body had been long dead, the letter of the infamous stamp killer, stamped onto the victims head with significant force was plain to see. Thirteen months and thirteen deaths, each one with a letter on their head, following the alphabet. Monroe only hoped they could stop the killer before they got to Z, or he would go postal.  & In a world where emotions were outlawed, Olivia discovered a hidden cache of antique stamps, each one imbued with a forbidden feeling. She carefully affixed a ``Love'' stamp to her forehead, savoring the warmth that spread through her body. Emboldened, she began secretly distributing the stamps to others, watching as they experienced joy, sorrow, and passion for the first time. Olivia's underground movement grew, with the stamps becoming a secret code, a letter of solidarity passed between rebels. But when the authorities discovered her illicit activities, Olivia had to make a choice: send her followers to safety while sacrificing herself, or let the flames of the resistance die out; she chose to be the stamp that would seal their freedom, even as it meant the end of her own. \\
         \midrule
         petrol, diesel, pump & During one night at my local gym I bumped into Vin Diesel who commented on my ability to lift my weight correctly. He decided to show me a better way to get my pump on. Whilst given the chance to ask him, I quipped if he had any siblings named after petrol. & The old pump, rusted and forgotten, stood alone in the overgrown field. Once a bustling hub for farmers filling their tractors with petrol and diesel, it now served as a silent sentinel to a bygone era. Sunflowers, their heads heavy with seeds, swayed around its base, their roots tapping into the forgotten fuel lines beneath the ground. A robin built its nest in the nozzle, chirping a cheerful melody that echoed through the quiet air. \\
         \midrule
         gloom, payment, exist & It wasn't all doom and gloom. Matt found many ways to exist without falling in despair. One small payment to a bartender in exchange for a pint of ale was one such way. & In the heart of the city, a peculiar antique shop existed, where time seemed to stand still. One rainy afternoon, a mysterious old man walked in, his eyes reflecting the gloom of a forgotten era. He handed the shopkeeper a small, ancient-looking orb, whispering, "This is the last payment for the debt I owe." The shopkeeper, a wise woman with knowing eyes, took the orb and placed it in a hidden chamber behind the counter. As the orb's light dimmed, the old man vanished, leaving behind a whisper of gratitude and a sense of timeless peace. \\
         \midrule
         organ, empire, comply & Did you hear the one about the unluckiest man in the world? He once fell off the Empire State Building but landed without breaking one single bone! A miracle - Some would even call it witchcraft! But that's where his luck ran out. You see, while he was lying on the ground, a stray dog ran up to him and bit him right in the eye! Now he is on the Organ donor list waiting for new corneas, however rumor is that he doesn't comply with their rules, so he may be known as one-eyed Pete forever! & Once upon a time, in a world where organs could speak, there existed a rebellious liver named Levi. Unlike his peers in the Digestive Empire, Levi refused to comply with the monotonous routine of breaking down substances. Instead, he dreamt of exploring the vast, uncharted territories of the human body. One fateful day, Levi's adventurous spirit led him through a tiny capillary, embarking on an epic journey that would change the body's harmony forever. His legend became a whispered tale among the cells, inspiring others to listen to their desires. \\
        \bottomrule
    \end{tabular}
    \caption{Most creative human and AI stories for each item set according to expert ratings.}
    \label{tab:most-creative}
\end{table*}

\begin{table*}[h]
    \centering
    \begin{tabular}{p{4cm} p{6cm} p{6cm}}
         \toprule
         \textbf{Item set} & \textbf{Human Story} & \textbf{AI Story} \\
         \midrule
         stamp, letter, send & I asked my friend about sending the letter as I was new in the country. He told me how to stamp the letter and and how can I send it by the post. I finally had a good understanding about it. & In a world where emotions are tangible, a young girl named Lila collected stamps of feelings, each one imbued with a unique sentiment. She carefully chose a ``Joy'' stamp, placed it on a letter she'd written to her grandmother, and with a whisper, sent it flying through the air, transforming the paper into a dove that carried her heartfelt message. \\
         \midrule
         petrol, diesel, pump & The man had quickly realised that he filled his tank with petrol instead of diesel when the engine cut out. Trying to pump it out of his tank was impossible as he had not tools to do so. He tried calling around to see if anyone had a means to fix his problem, but alas he has to call a local garage to pick him and his car up. He was hoping and praying that his car would be able to be fixed. It doesn't look good. & Once upon a time, in a small village nestled between two mountains, there lived a young boy named Tom. He was a curious and adventurous child, always exploring the world around him. One day, while wandering through the woods, he stumbled upon a hidden cave. Inside, he found a strange device made of old, rusty metal. \\
         \midrule
         gloom, payment, exist & In a far away place, a million miles away, gloom hung heavy in the air as if no one existed there at all. It was a quiet town, with minimal people who went about their business as if no one else was there. Happiness in the town felt like a distant memory for most, leaving only the weight of sorrow as payment. & In the gloom of the moonless night, an ancient tome fell open to a page inscribed with the arcane incantation for eternal life. Desperate and fearful, the reclusive alchemist agreed to make the forbidden payment, scribbling the ritual in blood onto a parchment of vellum. As the first rays of dawn crept through the shutters, he whispered the words of exist beyond the veil, unaware of the dark consequences awaiting him. \\
         \midrule
         organ, empire, comply & The British empire was at its peak, the general had reigning power on the poor locals. All they had was to comply to this force that had entered their country. The churches were erected and instilled their unjust values. Inside one church, was a vast organ that rung out and around the land. A reminder to the occupied. & In a world where music was outlawed, an underground organ empire flourished. A lone musician, skilled in the art of compliance, played a hidden organ, its haunting notes echoing through the silent streets, igniting the sparks of rebellion in the hearts of the oppressed. Unseen, they danced to the forbidden rhythm, their spirits rising in defiance against the empire's tyranny. And in the depths of the night, the organ sang, a beacon of hope and freedom, leading them to a future where music could once again fill the air. \\
        \bottomrule
    \end{tabular}
    \caption{Least creative human and AI stories for each item set according to expert ratings.}
    \label{tab:least-creative}
\end{table*}

\begin{figure*}[h]
\begin{subfigure}[b]{0.5\textwidth}
    \centering
    \includegraphics[width=\textwidth]{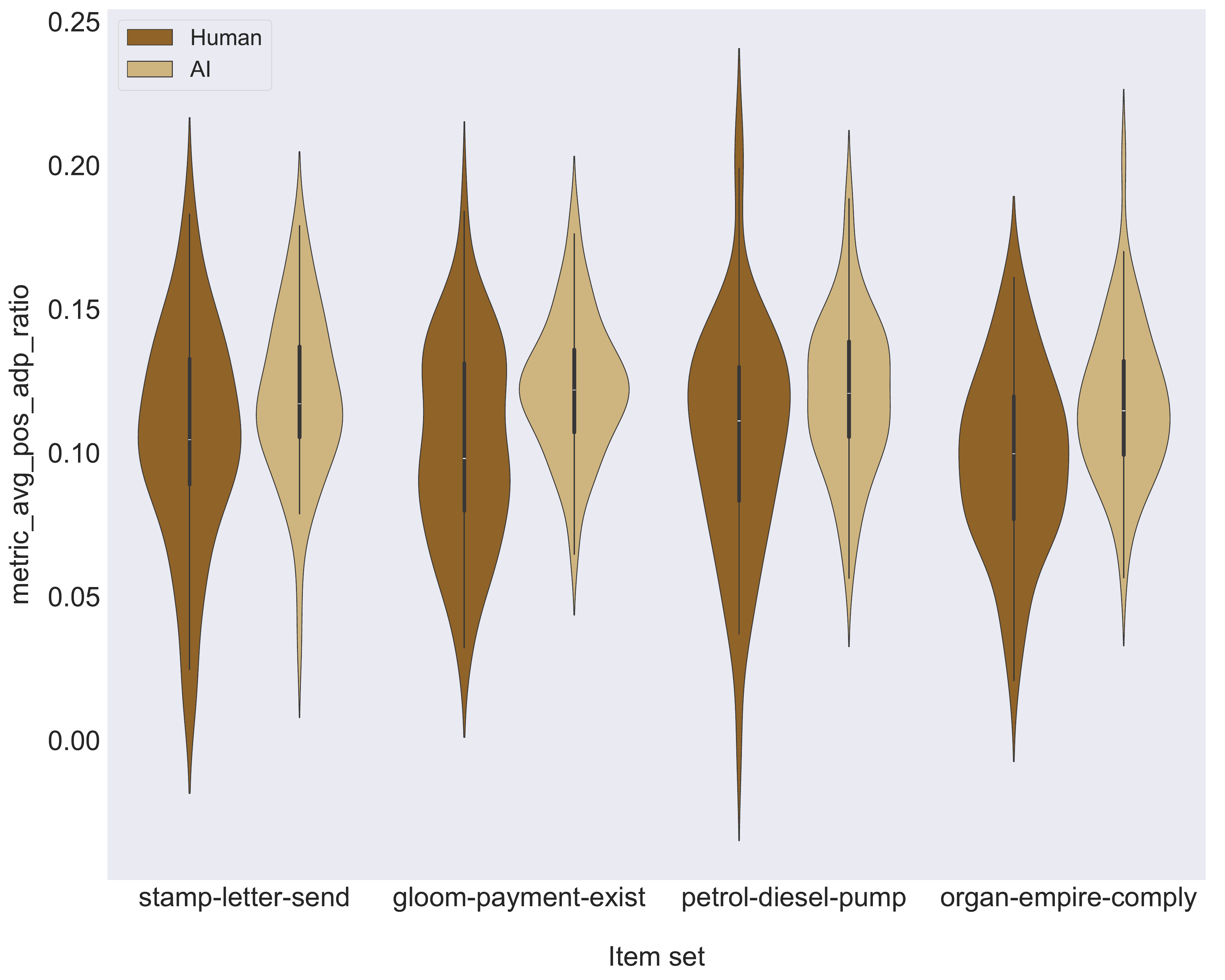}
    \caption{Average noun ratio per sentence.}
    \label{fig:results-noun-ratio}
\end{subfigure}
\begin{subfigure}[b]{0.5\textwidth}
    \centering
    \includegraphics[width=\textwidth]{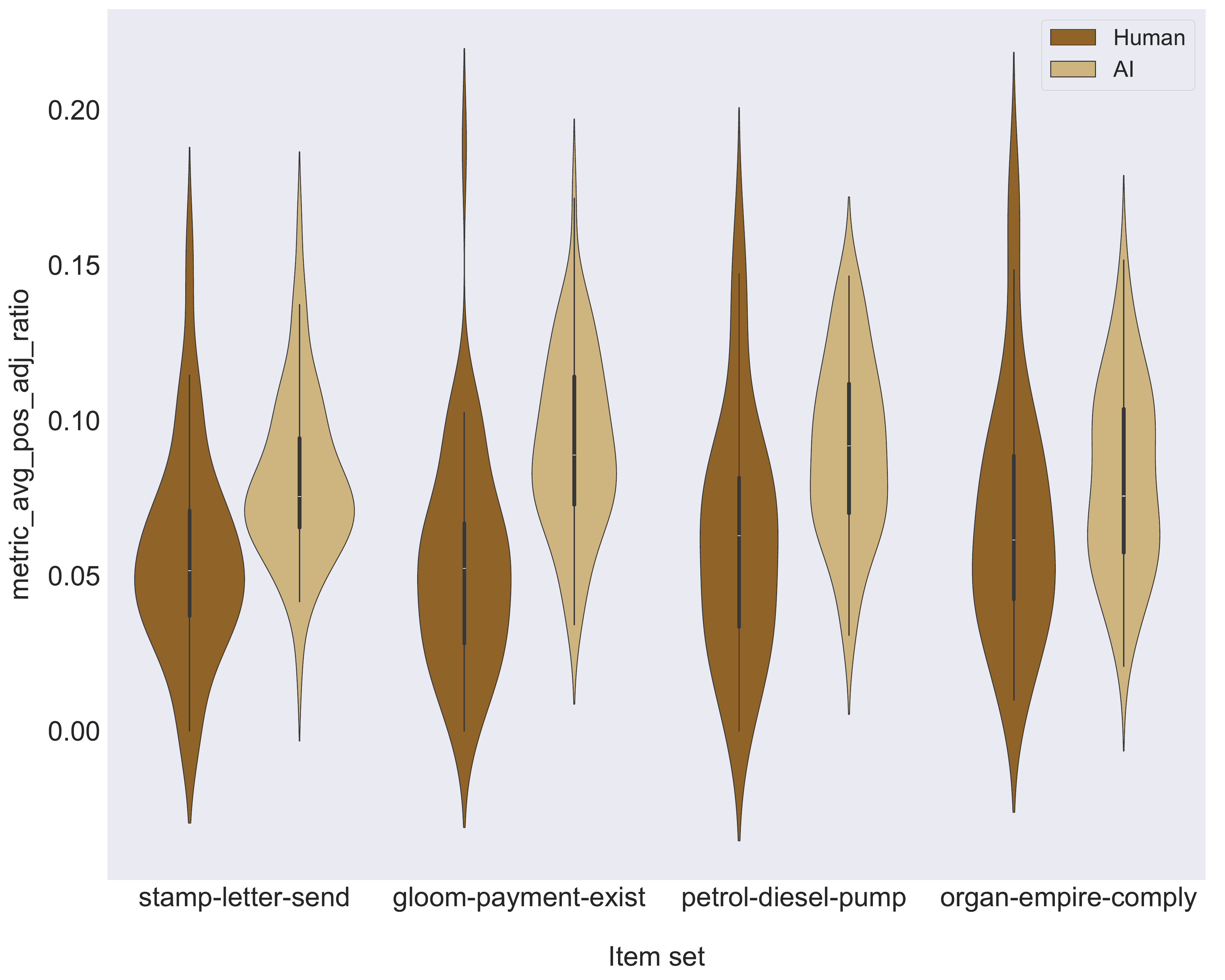}
    \caption{Average adjective ratio per sentence.}
    \label{fig:results-adj-ratio}
\end{subfigure}
\begin{subfigure}[b]{0.5\textwidth}
    \centering
    \includegraphics[width=\textwidth]{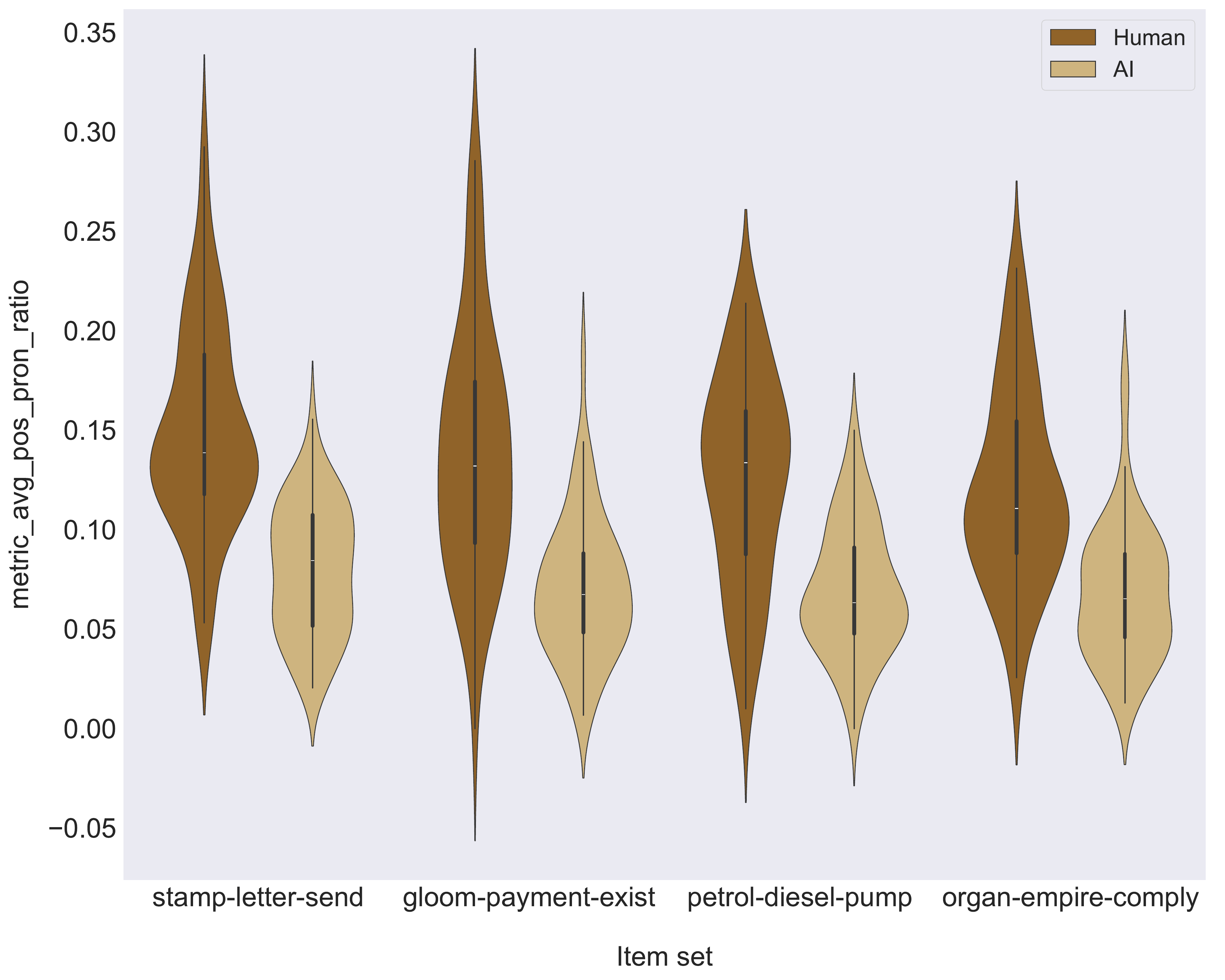}
    \caption{Average pronoun ratio per sentence.}
    \label{fig:results-pronoun-ratio}
\end{subfigure}
\begin{subfigure}[b]{0.5\textwidth}
    \centering
    \includegraphics[width=\textwidth]{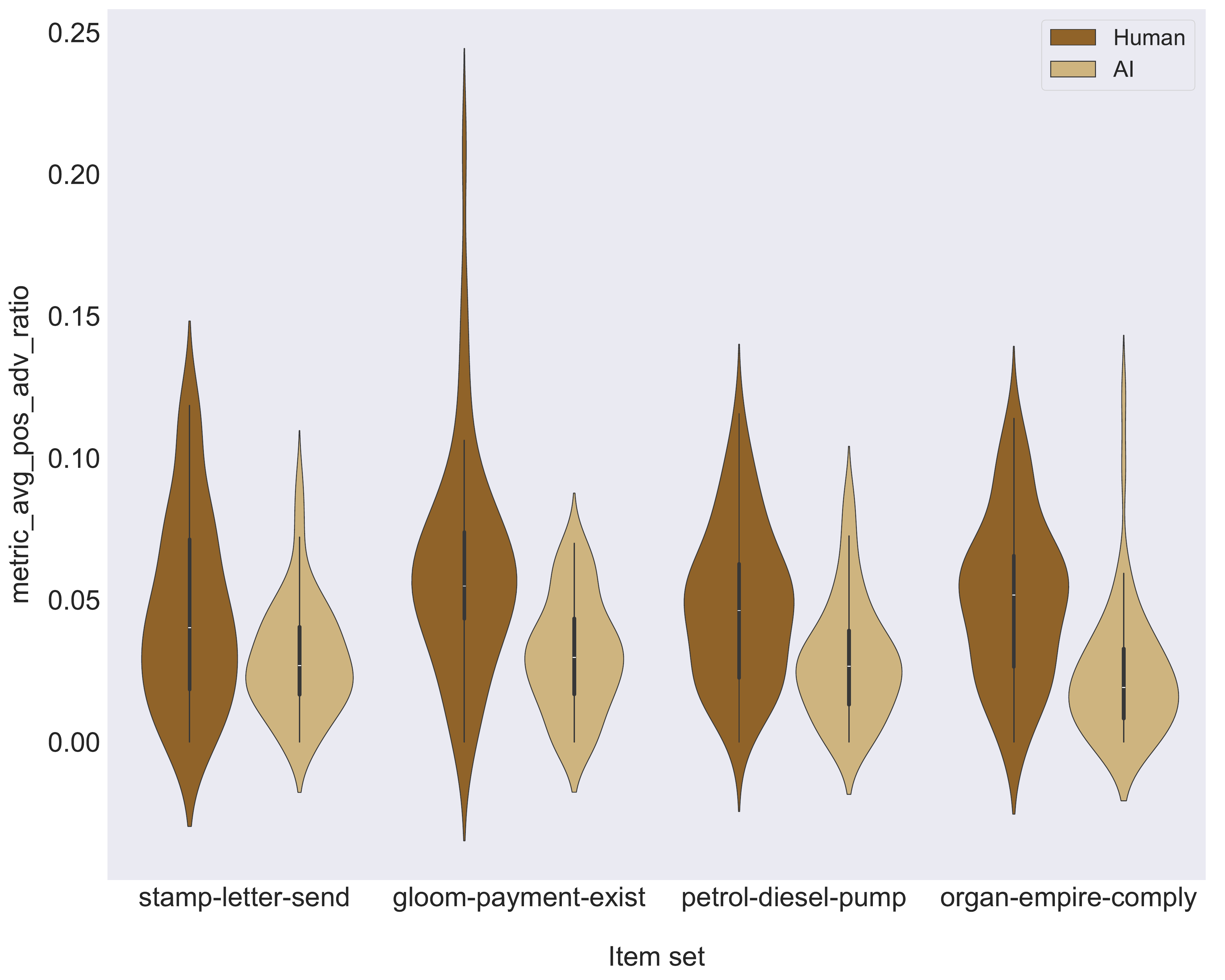}
    \caption{Average adverb ratio per sentence.}
    \label{fig:results-adv-ratio}
\end{subfigure}
\caption{Syntactic complexity scores measured by average POS tag ratios.}
\label{fig:results-pos-ratio}
\end{figure*}

\begin{figure*}[h]
\begin{subfigure}[b]{0.5\textwidth}
    \centering
    \includegraphics[width=\textwidth]{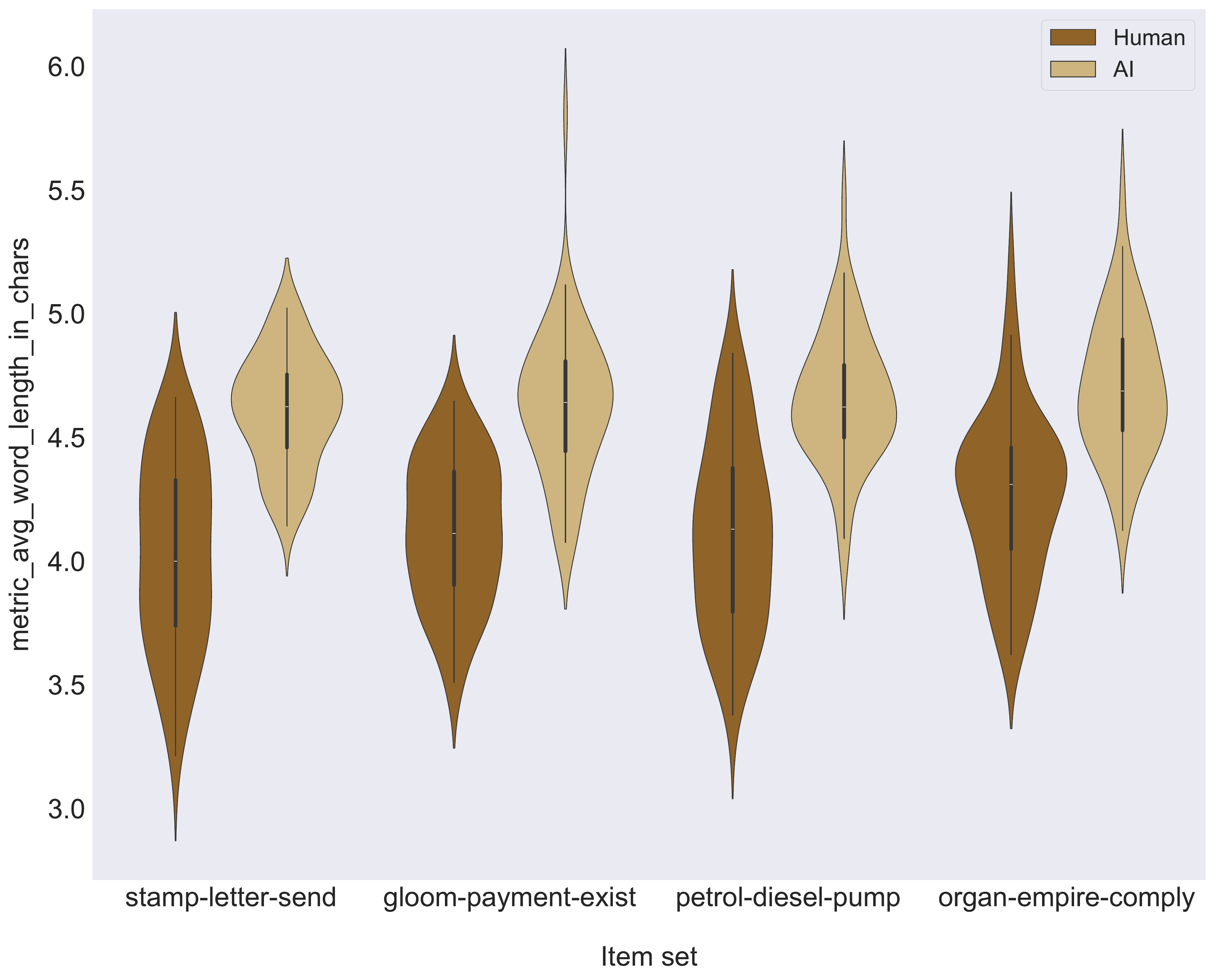}
    \caption{Average word length in number of characters.}
    \label{fig:results-word-length-in-chars}
\end{subfigure}
\begin{subfigure}[b]{0.5\textwidth}
    \centering
    \includegraphics[width=\textwidth]{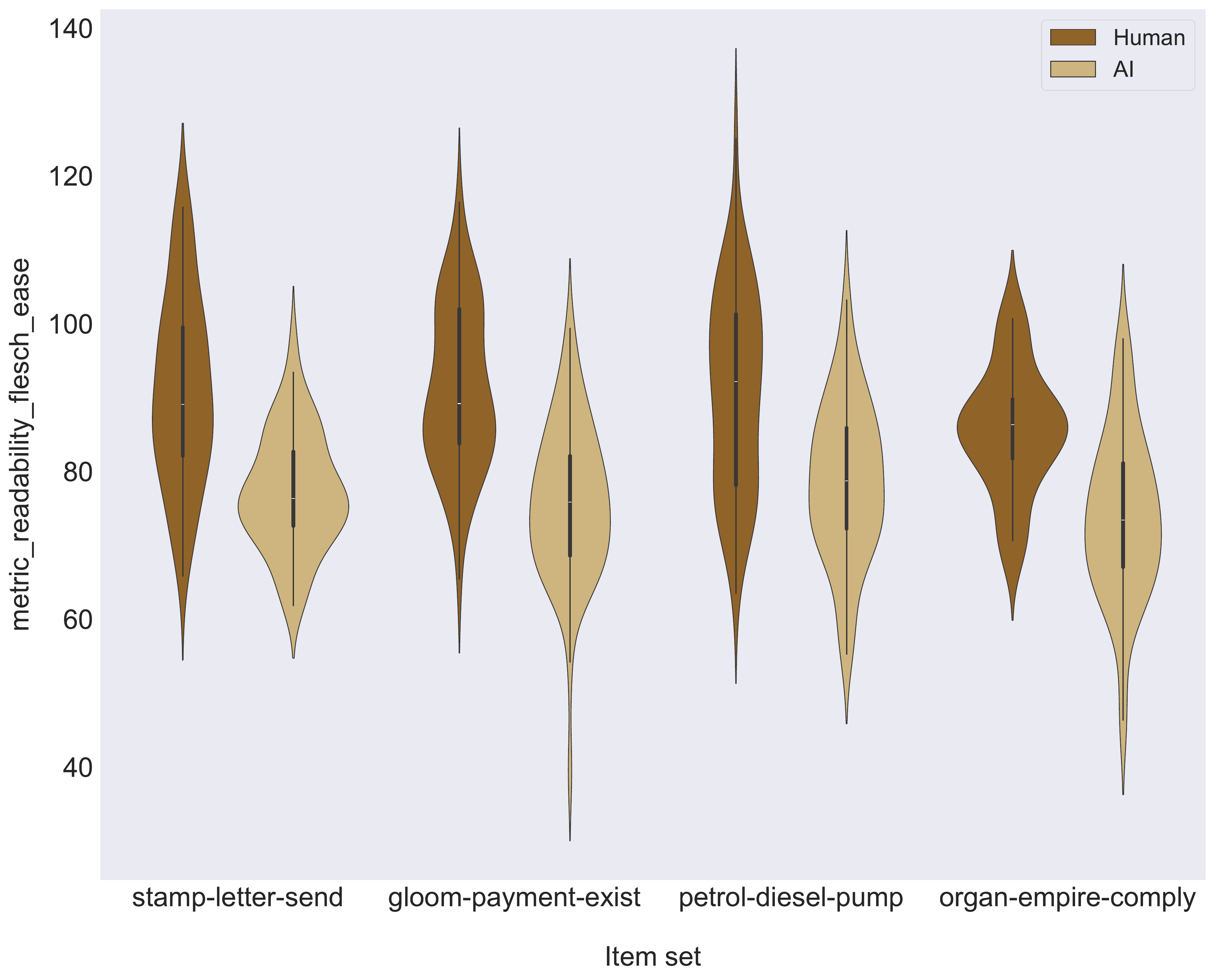}
\caption{Flesch reading ease scores. Higher means easy readability.}
\label{fig:results-readability}
\end{subfigure}
\begin{subfigure}[b]{0.5\textwidth}
    \centering
    \includegraphics[width=\textwidth]{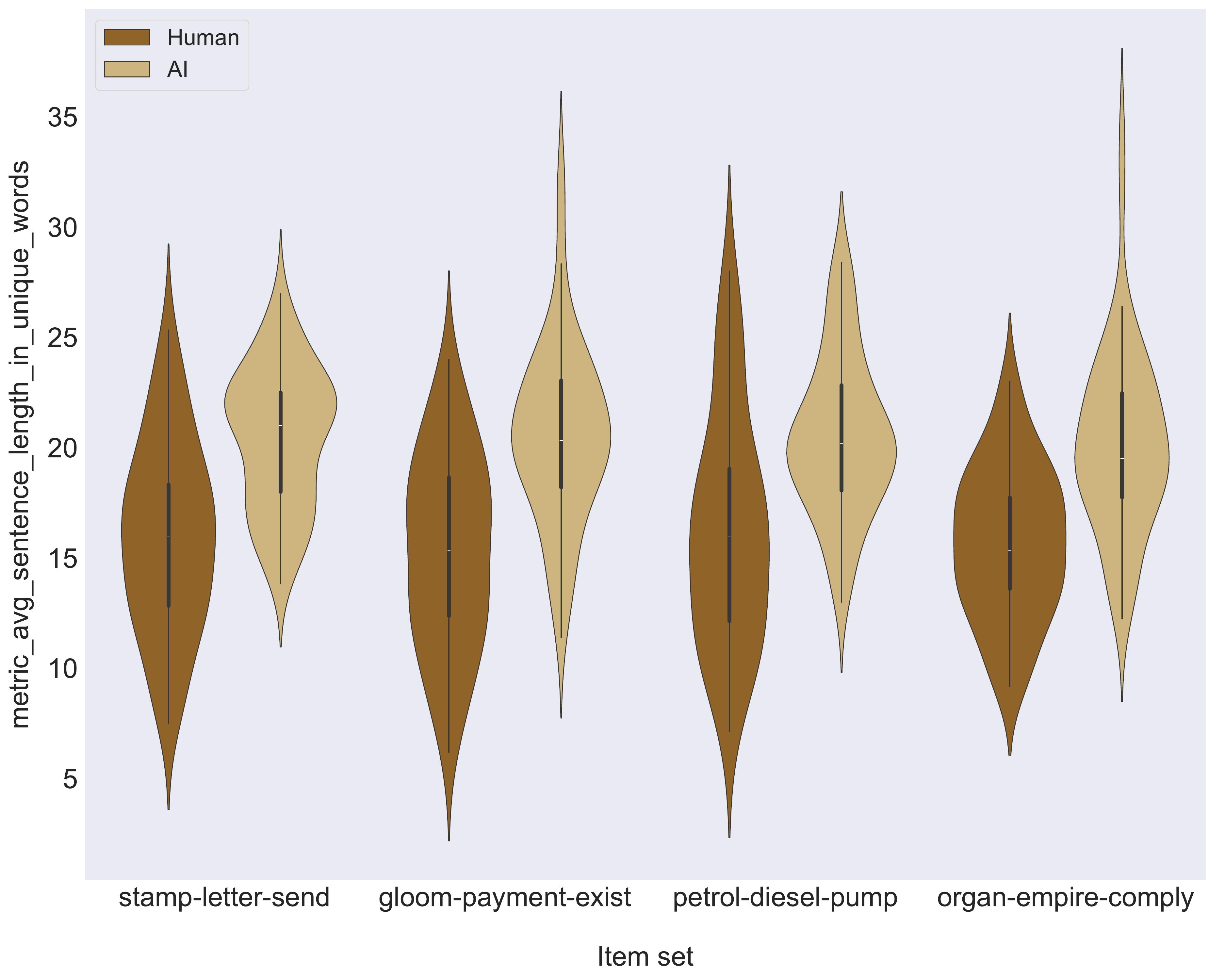}
    \caption{Average sentence length in number of unique words.}
    \label{fig:results-sent-length-in-words}
\end{subfigure}
\begin{subfigure}[b]{0.5\textwidth}
    \centering
    \includegraphics[width=\textwidth]{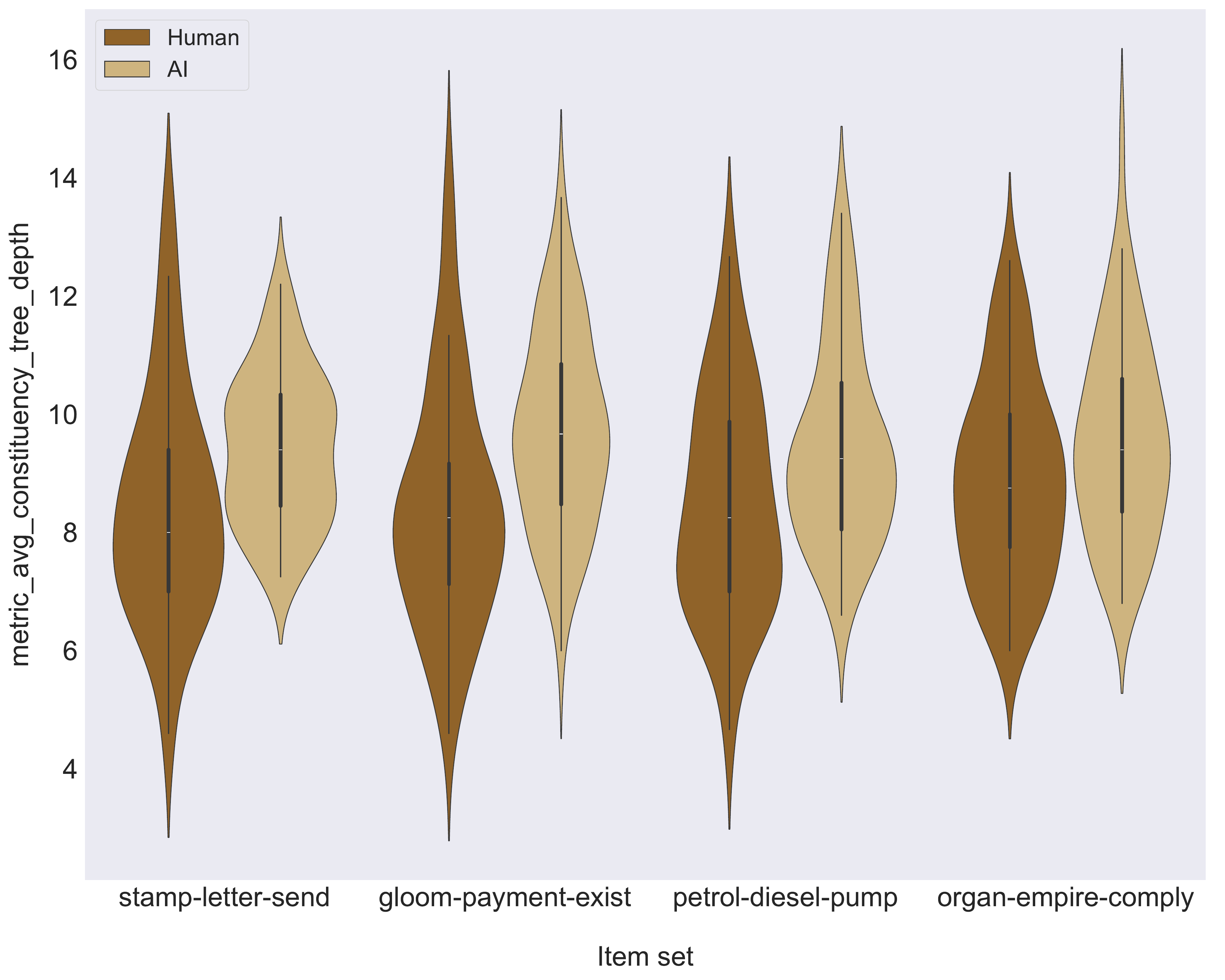}
    \caption{Average constituency tree depth.}
    \label{fig:results-avg-const-tree-depth}
\end{subfigure}
\caption{Results for additional lexical and syntactic complexity metrics.}
\label{fig:results-more-complexity}
\end{figure*}

\begin{figure*}[h]
\begin{subfigure}[b]{0.5\textwidth}
    \centering
    \includegraphics[width=\textwidth]{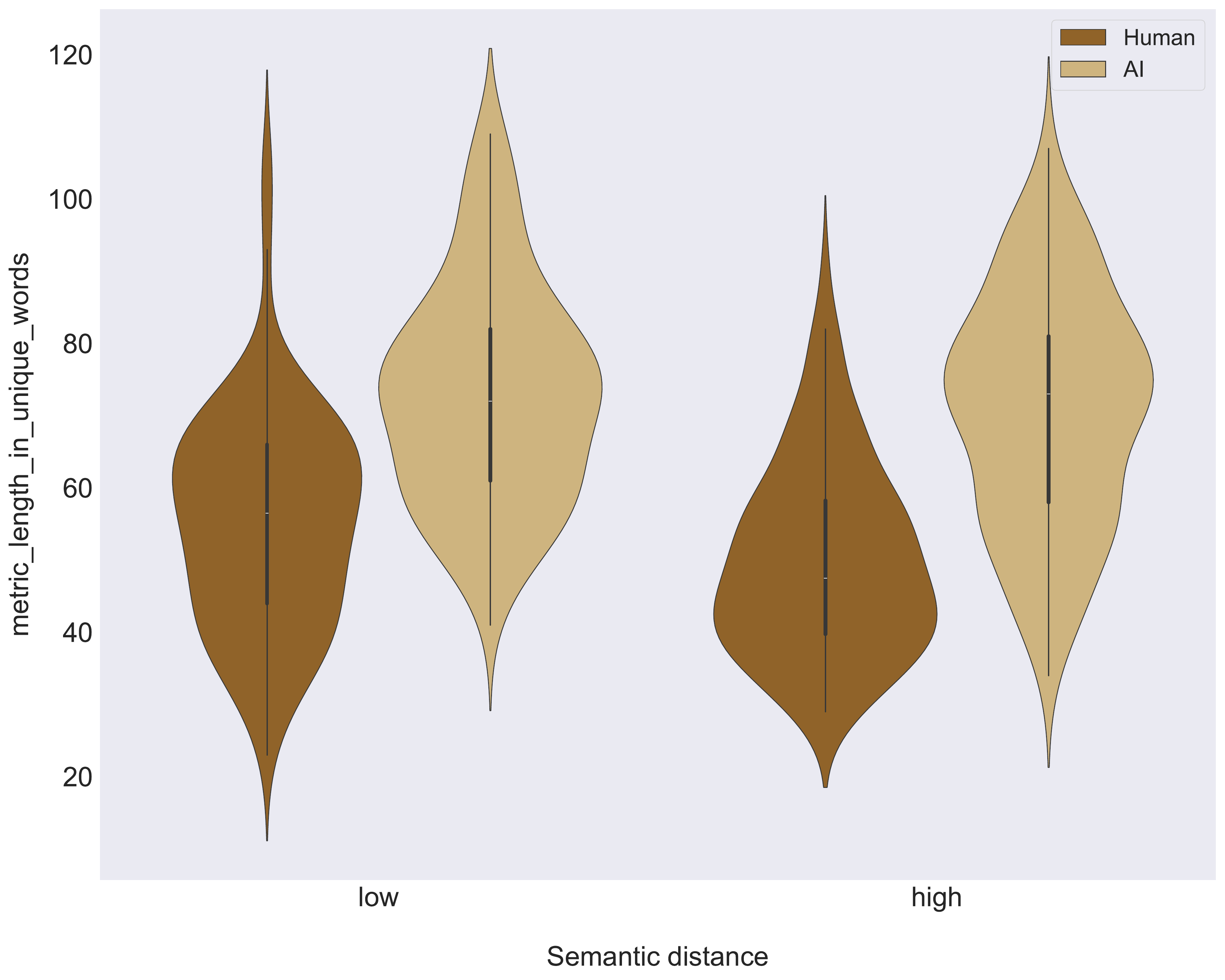}
    \caption{Lexical complexity scores stratified by semantic distance measured by story length in number of unique words.}
    \label{fig:results-semdis-length-in-words}
\end{subfigure}
\begin{subfigure}[b]{0.5\textwidth}
    \centering
    \includegraphics[width=\textwidth]{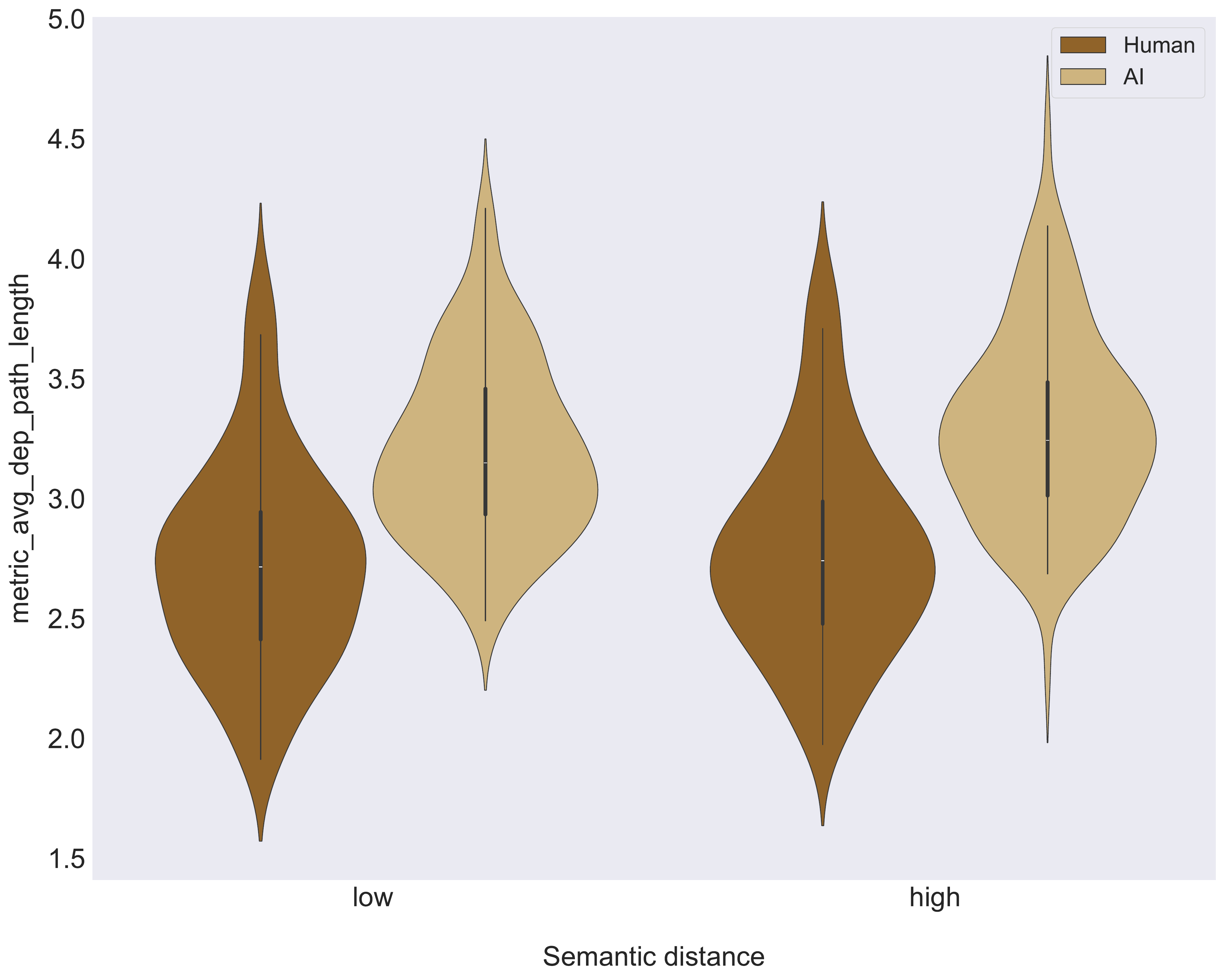}
\caption{Syntactic complexity scores stratified by semantic distance measured by average dependency path length.}
\label{fig:results-semdis-avg-dep-path}
\end{subfigure}
\caption{Results for lexical and syntactic complexity metrics stratified by semantic distance.}
\label{fig:results-semdis-complexity}
\end{figure*}

\begin{figure*}[h]
\begin{subfigure}[b]{0.5\textwidth}
    \centering
    \includegraphics[width=\textwidth]{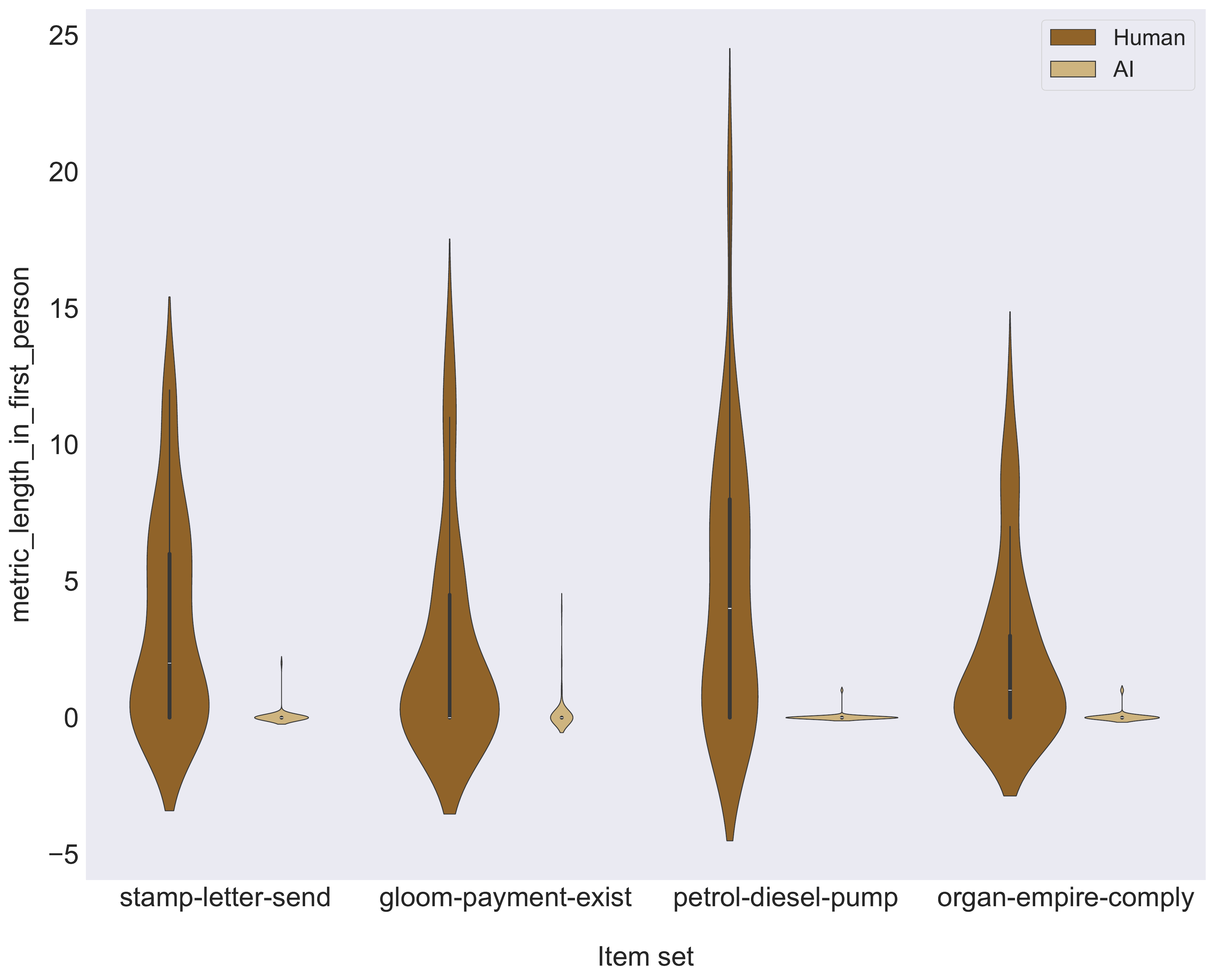}
    \caption{First person pronoun use.}
    \label{fig:results-first-person}
\end{subfigure}
\begin{subfigure}[b]{0.5\textwidth}
    \centering
    \includegraphics[width=\textwidth]{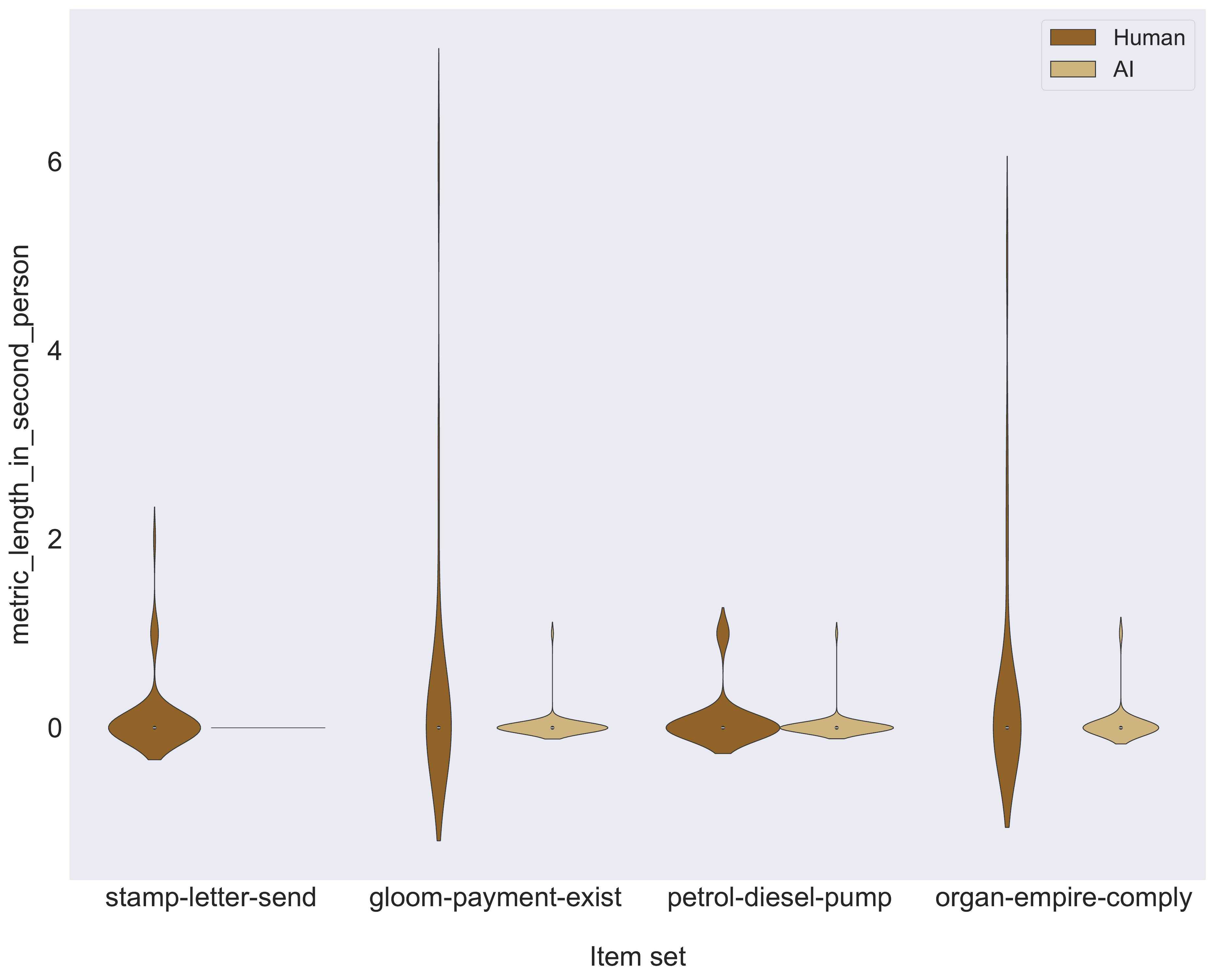}
    \caption{Second person pronoun use.}
    \label{fig:results-second-person}
\end{subfigure}
\begin{subfigure}[b]{0.5\textwidth}
    \centering
    \includegraphics[width=\textwidth]{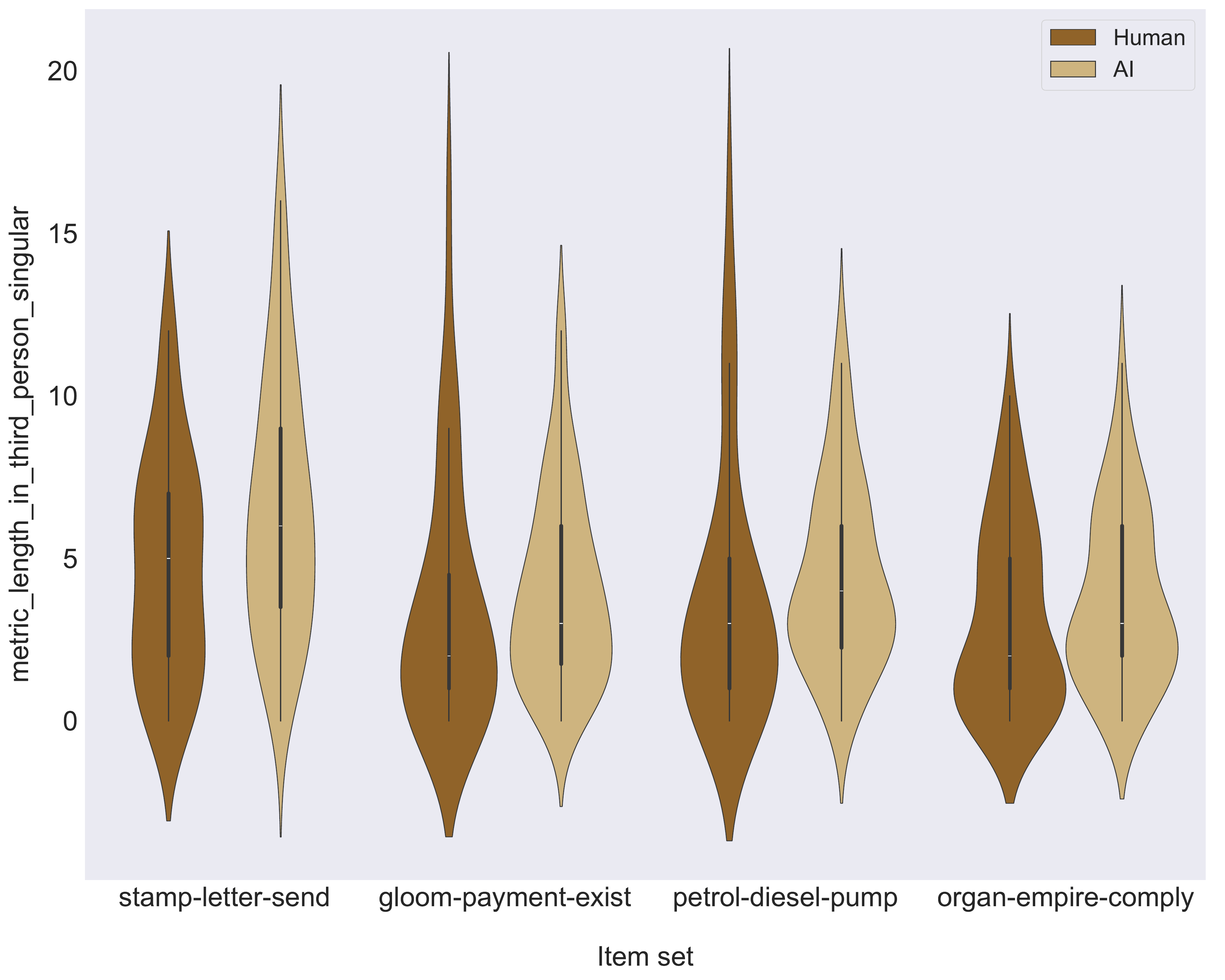}
    \caption{Third person singular pronoun use.}
    \label{fig:results-third-person-sing}
\end{subfigure}
\begin{subfigure}[b]{0.5\textwidth}
    \centering
    \includegraphics[width=\textwidth]{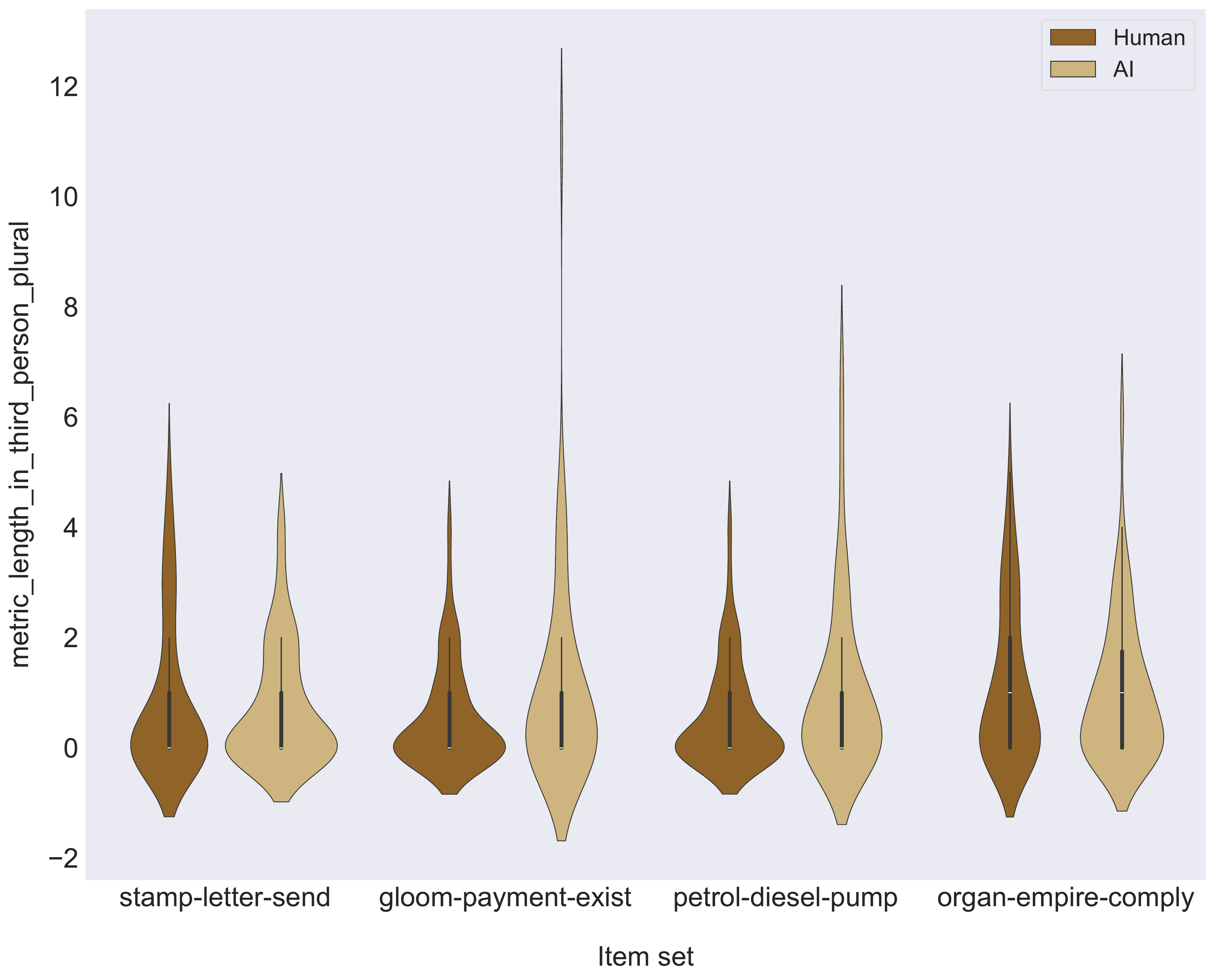}
    \caption{Third person plural pronoun use.}
    \label{fig:results-third-person-pl}
\end{subfigure}
\caption{Results for pronoun use analysis.}
\label{fig:results-pronoun-use}
\end{figure*}

\begin{figure*}[h]
\begin{subfigure}[b]{0.5\textwidth}
    \centering
    \includegraphics[width=\textwidth]{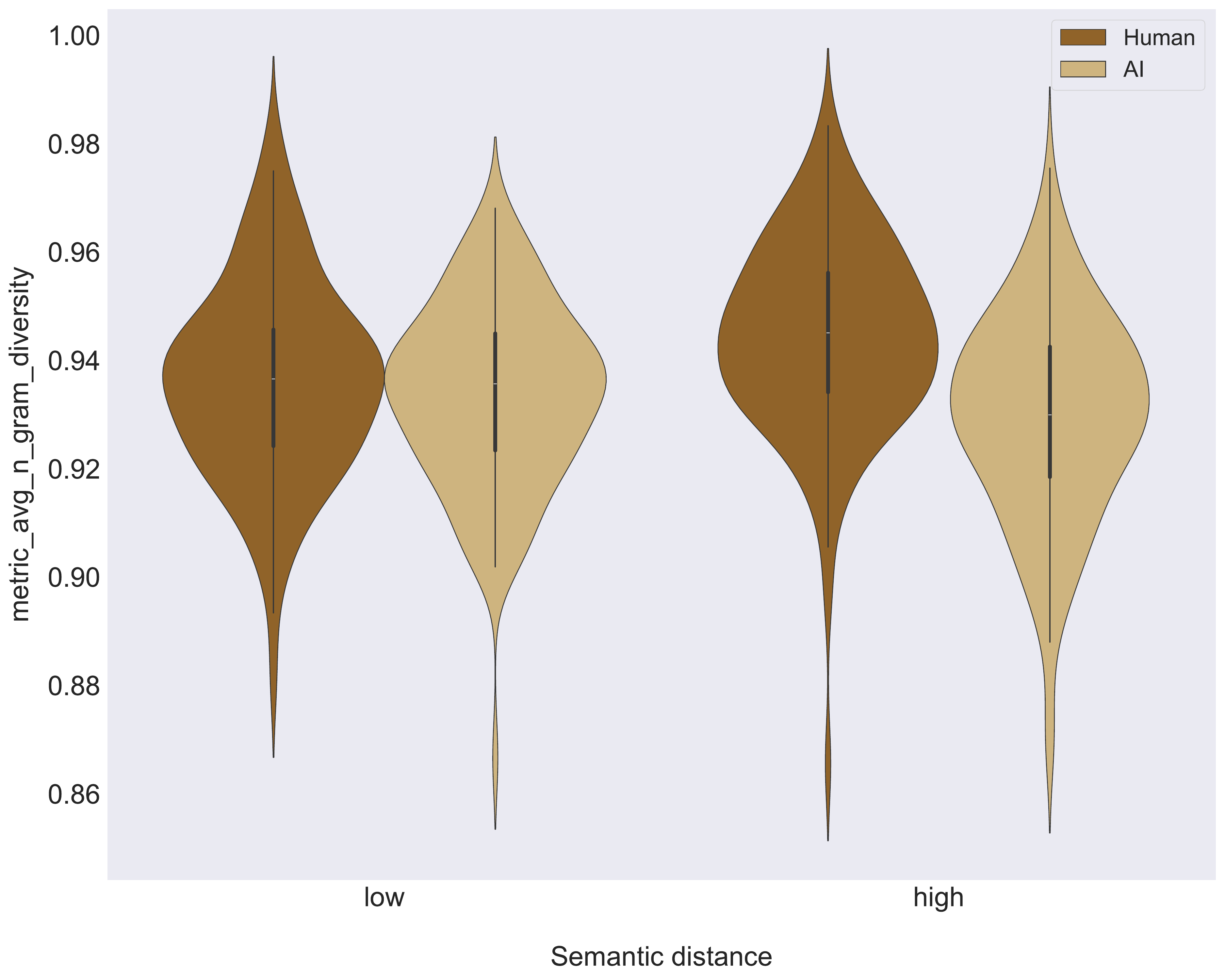}
    \caption{\textit{n}-gram diversity scores stratified by semantic distance.}
    \label{fig:results-n-gram-diversity-semdis}
\end{subfigure}
\begin{subfigure}[b]{0.5\textwidth}
    \centering
    \includegraphics[width=\textwidth]{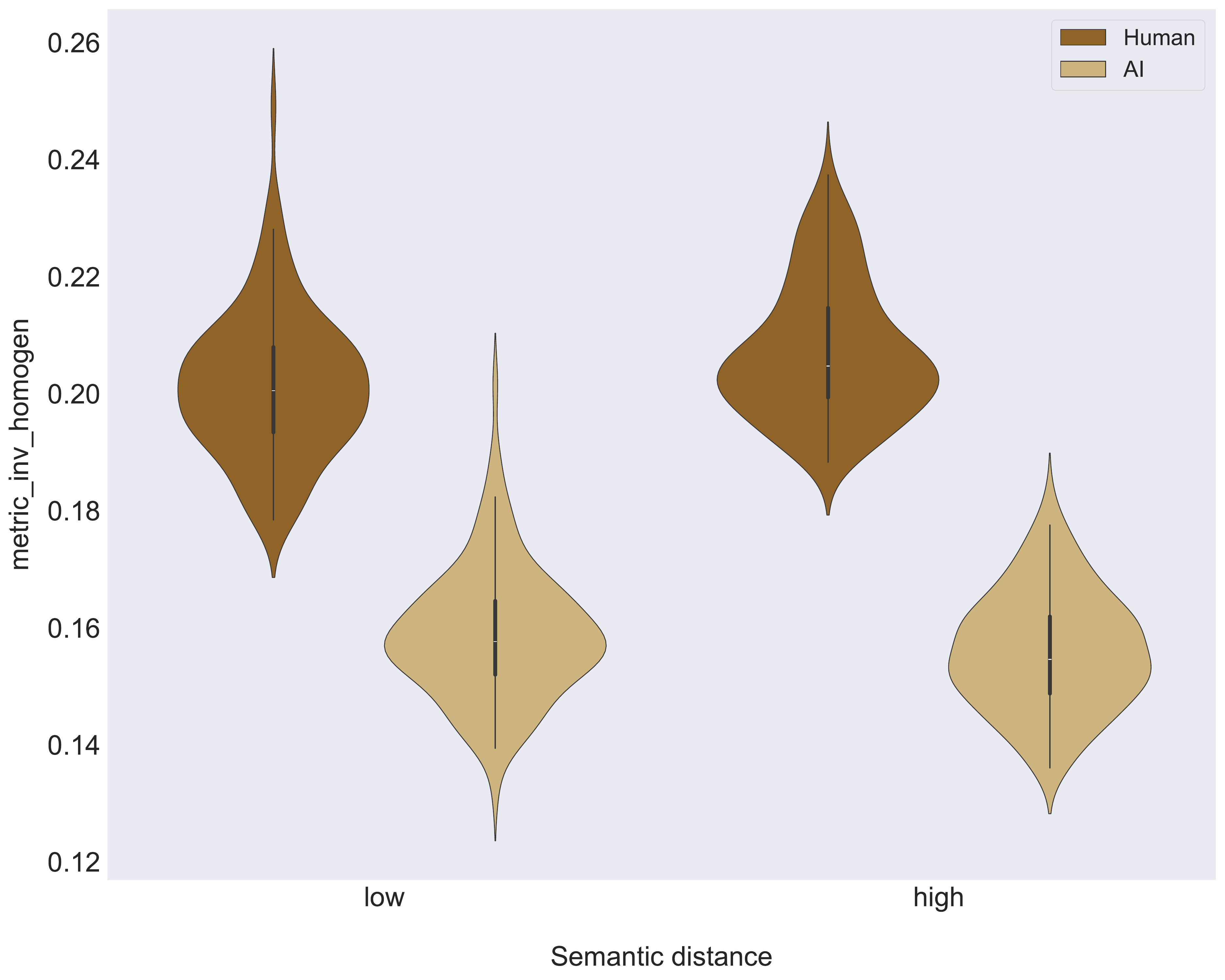}
    \caption{Inverse homogenization scores stratified by semantic distance.}
    \label{fig:results-inv-homogen-semdis}
\end{subfigure}
\caption{\textit{n}-gram diversity and inverse homogenization scores stratified by semantic distance between cue words.}
\label{fig:results-diversity-semdis}
\end{figure*}

\begin{figure*}[h]
\begin{subfigure}[b]{0.5\textwidth}
    \centering
    \includegraphics[width=\textwidth]{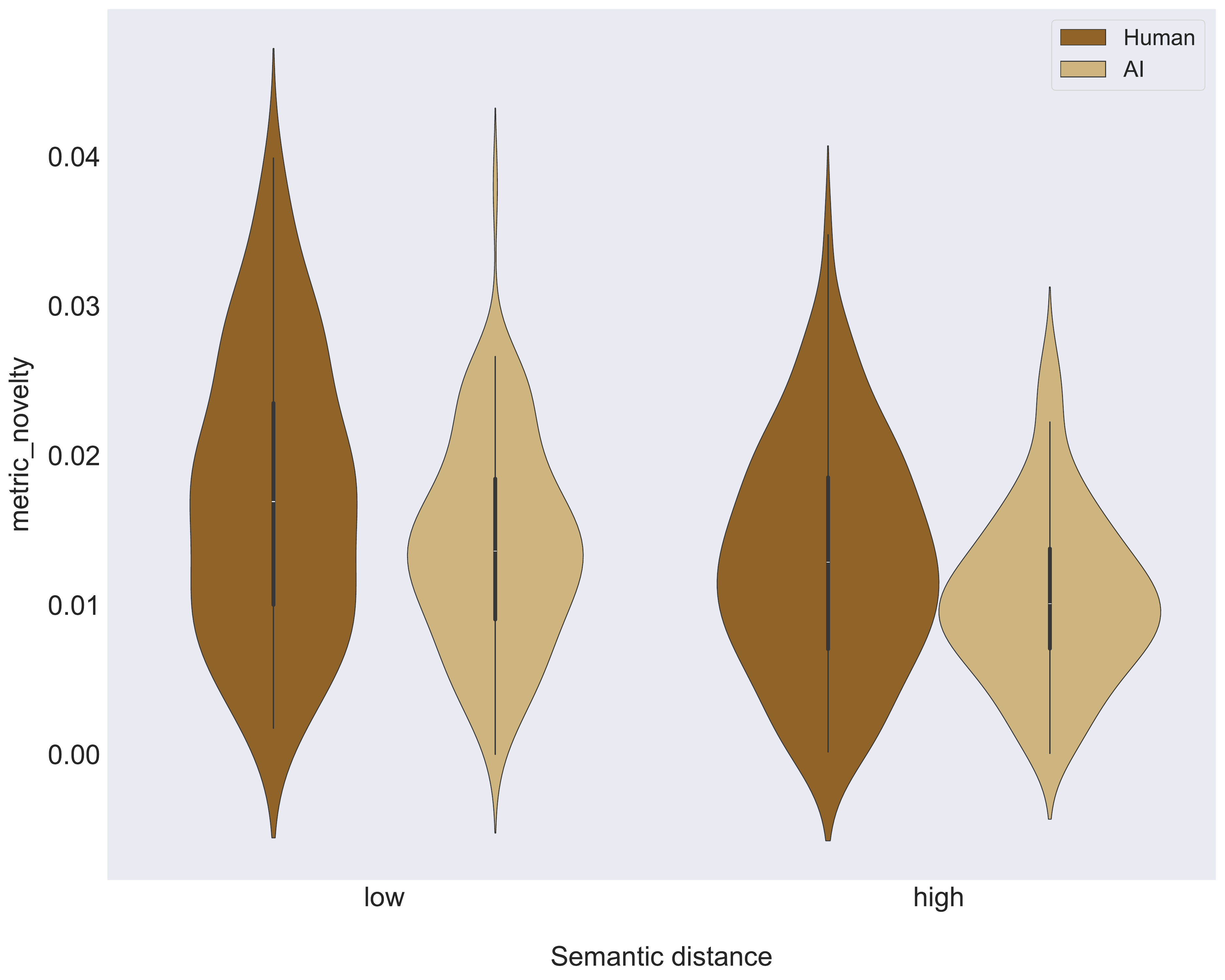}
    \caption{Novelty scores stratified by semantic distance.}
    \label{fig:results-novelty-semdis}
\end{subfigure}
\begin{subfigure}[b]{0.5\textwidth}
    \centering
    \includegraphics[width=\textwidth]{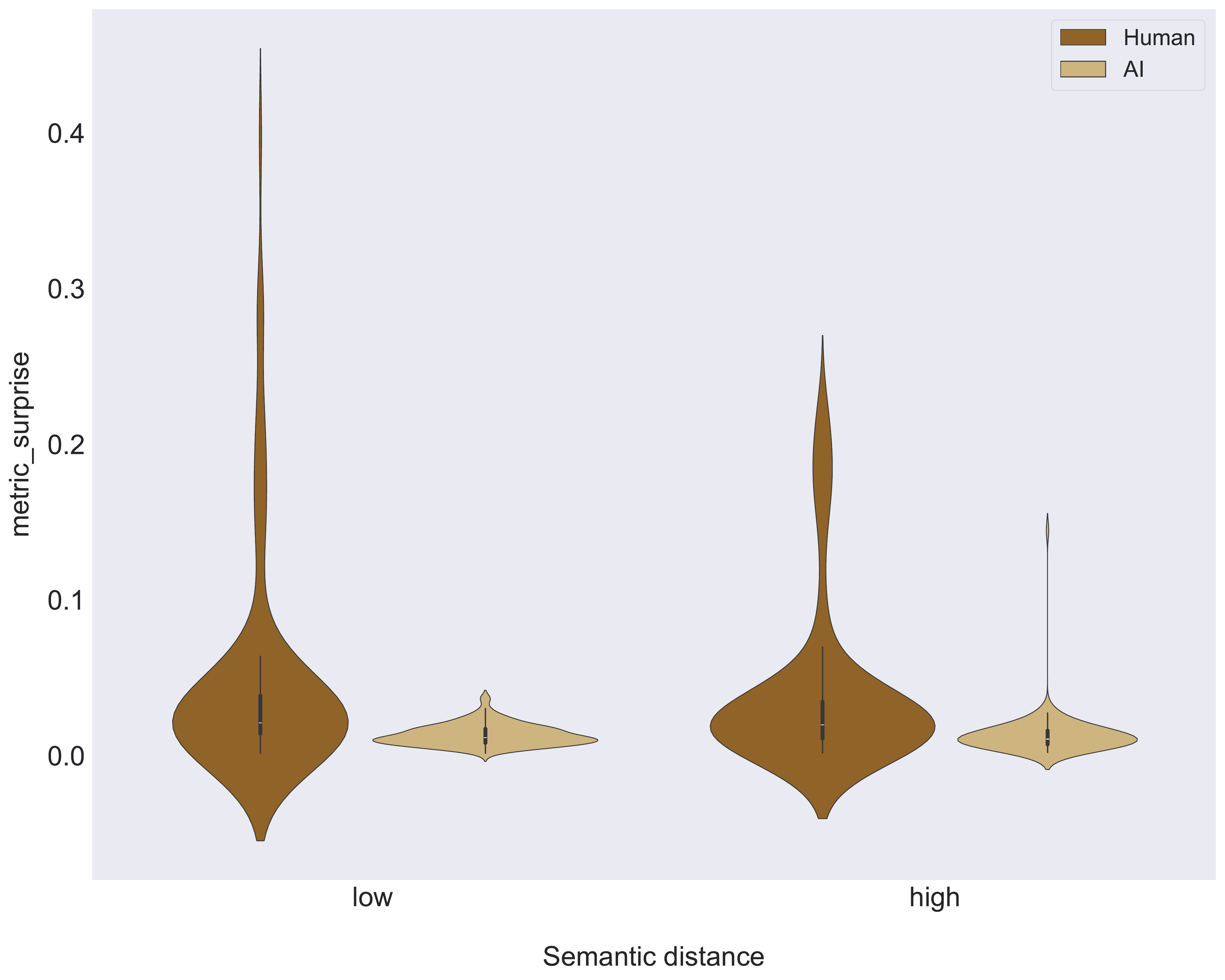}
    \caption{Surprise scores stratified by semantic distance.}
    \label{fig:results-surprise-semdis}
\end{subfigure}
\caption{Novelty and surprise scores stratified by semantic distance between cue words.}
\label{fig:results-novelty-surprise-semdis}
\end{figure*}

\begin{table}
\centering
\begin{tabular}{ llll }
\textbf{Item set} & \textbf{Human Stories} & \textbf{AI Stories} & \textbf{Total} \\
 \toprule
 stamp, letter, send & 53 & 55 & 108 \\
 petrol, diesel, pump & 51 & 58 & 109\\
 organ, empire, comply & 53 & 54 & 107\\
 gloom, payment, exist & 51 & 56 & 108 \\
 \midrule
 Total & 208 & 223 & 431 \\
 \bottomrule
\end{tabular}
\caption{Final data statistics for each item set and evaluation group.}
\label{tab:final-data-stats}
\end{table}